\documentclass{article} %
 \pdfpageattr{/Group <</S /Transparency /I true /CS /DeviceRGB>>}
        \usepackage{iclr2027_conference,times}
        \usepackage{graphicx}
        \usepackage{iclr2027_conference,times}
        
        \usepackage{xcolor}
        \usepackage{hyperref}
        \usepackage{url}
        \usepackage{graphicx}
        \usepackage{subcaption}
        \usepackage{booktabs}
        \usepackage{multirow}
        \usepackage{amsmath}
        \usepackage{amsfonts}
        \usepackage{footnote}
        \usepackage{fontawesome5}
        
        \usepackage{tipa}
        \newcommand{\ipa}[1]{\textipa{#1}}
        \newcommand{\rhotic}{[\textipa{\*r}]}      %
        \newcommand{\uvular}{[\textipa{K}]}        %
        
        \newcommand{\morris}[1]{}
        \newcommand{\moran}[1]{}
        \newcommand{\todo}[1]{}
        
        \newcommand{\powsm}{POWSM}
        \newcommand{\pxeus}{PhoneticXeus}

        \usepackage[normalem]{ulem} %
        \usepackage{xcolor}
        
        \usepackage{lipsum}
        \usepackage{amsfonts}
        \usepackage{amsmath,amssymb}

        \title{\centering Phonological Interference \\ in Multilingual Speech Models}
        
        \author{\hspace*{-\tabcolsep}\parbox[t]{\textwidth}{\centering
        Moran Yanuka$^{1}$ \qquad Raja Giryes$^{1}$ \qquad Morris Alper$^{2}$ \\[4pt]
        {\mdseries $^{1}$Tel Aviv University \qquad $^{2}$University of Miami} \\[6pt]
        {\mdseries\small \faGlobe\ \href{https://moranyanuka.github.io/phonological-interference/}{moranyanuka.github.io/phonological-interference}}
        }}

        \iclrfinalcopy %
        \begin{document}
        
        \maketitle
        \lhead{Preprint. Under review.} %
        
        \begin{abstract}
        Phoneme-level models transcribe or generate speech as a sequence of phonemes, the smallest sound units that distinguish words. These models enable fine-grained pronunciation control and understanding, yet often fail on input that does not match any single training language, such as speech alternating between two languages, known as code-switching, or low-resource languages absent from training. We identify a systematic failure mode behind this, \emph{phonological interference}: models assume the input is in a single language and impose its phonology, overriding local phoneme-level decisions that conflict with the assumed language. We measure interference by how often a model retains phonemes that one language has but the other lacks. On code-switched input, two phone recognizers (speech-to-phoneme models) and a phoneme-conditioned text-to-speech model lose 32\% to 79\% of these phonemes, but lose far fewer of the phonemes both languages share. On unseen languages, we find that phone recognizers impose the phonology of the training language they assign to the speech, and the more confident the assignment, the more they lose phonemes the unseen language has but the assigned language lacks. We probe the models' language estimate from their internal activations, and trace interference to a low dimensional subspace. On monolingual speech, steering this subspace toward another language makes the model lose the phonemes that only the original language uses and produce phonemes that only the target language has.
        We introduce \emph{windowed language estimation} (WLE), an inference time repair that replaces the model's language estimate in this subspace with one computed from a short window around each position. On code-switched input, WLE removes 34\% to 69\% of the interference in all three models, and in the recognizers it leaves monolingual performance essentially unchanged.
        \end{abstract}

        \section{Introduction}
        \label{sec:intro}

Phoneme-level speech models operate on phonemes, the smallest sound units that distinguish words, rather than words or characters. This makes them useful wherever pronunciation is important: documenting low-resource languages, assessing non-native speech, and precise speech synthesis. Yet they often fail on precisely the input they are meant to handle: speech that matches no single training language. This happens when speakers switch languages mid-sentence (code-switching), common in bilingual communities~\citep{bullock2009cambridge}, or when transcribing languages unseen during training, a central use of universal phone recognizers~\citep{li2020universal}.

\begin{figure}[t]
            
            \centering
            
            \begin{subfigure}[b]{0.485\linewidth}
                \includegraphics[width=\linewidth]{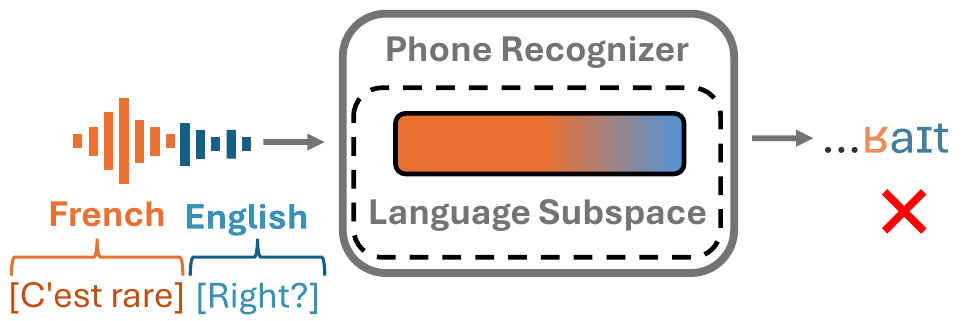}
                \caption{Phonological interference}
                \label{fig:overview-a}
            \end{subfigure}\hfill
            {\color{black!30}\vrule width 0.6pt}\hfill
            \begin{subfigure}[b]{0.46\linewidth}
                \includegraphics[width=\linewidth]{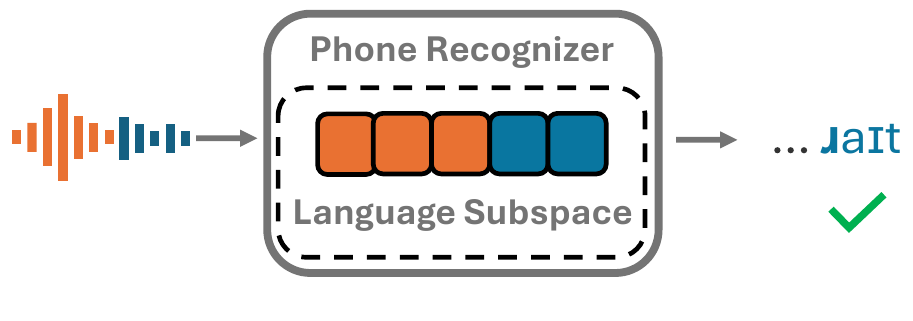}
                \caption{Windowed language estimation (WLE)}
                \label{fig:overview-b}
            \end{subfigure}
          \caption{
          \textbf{Phonological interference.} Multilingual phoneme-level speech models estimate the language from the whole input and override individual phoneme predictions.
          (a)~In a French--English code-switched utterance, a phone recognizer replaces English \rhotic{} in \textit{right} with French \uvular{}. The bar shows the model's language estimate in a low dimensional language subspace at each position, with time running left to right. The model computes this estimate from the whole utterance, so it blends French (orange) into English (blue), and the model still reads the English span partly as French.
          (b)~Windowed language estimation (WLE) replaces the language subspace coordinates at each position with those the model computes from a short window of audio around it. The estimate is then computed from the local language context, and \rhotic{} is correctly transcribed.
          }
           
            \label{fig:overview}
              \vspace{-1.em}
        \end{figure}

We identify a systematic pattern underlying this failure, which we term \emph{phonological interference}: the model contextually infers which language is being processed and imposes that language's phonology (sound system), overriding local decisions about individual phonemes.
This term is inspired by similar interference effects in studies of human bilingualism~\citep{weinreich1953languages}.
As illustrated in Figure~\ref{fig:overview}, while a model trained on monolingual English and French may pronounce each language correctly in isolation, it conflates phonemes that are distinct across the two languages in mixed utterances, such as the French and English ``r'' sounds, represented by \uvular{} and \rhotic{}, respectively, in the International Phonetic Alphabet (IPA)~\citep{international1999handbook}.

This failure mode is a manifestation of \emph{language confusion}, a phenomenon documented across modalities: multilingual speech models struggle with code-switched input~\citep{liu2026code, lee2025unicom}, while text LLMs have been observed to generate text in the wrong language~\citep{marchisio2024understanding, park2026code}. We extend this line of work to phoneme-level speech models, where we find that language confusion manifests as systematic substitution of individual phonemes. We demonstrate this phenomenon in both phone recognition (speech-to-phonemes) and synthesis (phonemes-to-speech), and provide an account of it by tracing these substitutions to a specific internal representation that causally steers them.

We measure interference with an evaluation based on \emph{unshared phoneme recall}: how often a model preserves phonemes that occur in one language but not the other. The relative drop in phonemes which are unshared between the two languages compared to the phonemes that are shared between them separates interference from a general drop in accuracy, which would lower both alike. We evaluate two phone recognizers and a phoneme-conditioned TTS model on utterances that switch between two training languages. We further evaluate the recognizers on monolingual speech in 43 languages absent from training. In these languages, the recognizers infer one of their training languages and impose its phonology, causing interference even in the absence of code-switching.

We trace interference to a low dimensional language subspace in the model's representations, analogous to language subspaces documented in text LLMs~\citep{chang2022geometry, xie2022discovering, dumas2025separating}. Steering this subspace toward another language on monolingual speech makes the model drop the phonemes that only the original language uses and replace them with phonemes of the target language, while keeping most of the phonemes the two languages share. 
We find an analogous subspace in Whisper, a standard ASR model, and show that editing it changes the language of the output transcript, linking phonological interference to language confusion in speech recognition.

Our analysis suggests an inference-time repair. If interference comes from a language estimate the model forms over the whole input, replacing it with a local estimate should let each language span keep its own phonology. We call this \emph{windowed language estimation} (WLE): at each position, we replace the hidden state's coordinates in the language subspace with those the model computes from a short window of input around that position, leaving all other coordinates unchanged. Because this local estimate comes from the model's own representation of the same input, WLE needs no language label or external language identifier. It reduces phonological interference by 34\%-69\% across all three models and leaves the phone recognizers' monolingual error rates essentially unchanged.

Our contributions are: (1) we identify phonological interference and measure it in phone recognition and synthesis, on code-switched speech and unseen languages (Section~\ref{sec:bleed-results}); (2) we trace it to a low dimensional subspace that causally steers the output in phone recognizers, a TTS model and standard ASR (Section~\ref{sec:mechanism}); (3) we introduce WLE, an inference-time repair that takes its target from the model's own estimate on a short window and reduces interference in all three models (Section~\ref{sec:repair}).

    \section{Measuring Phonological Interference}
    \label{sec:setup}
    \label{sec:bleed-results}

    \paragraph{Definition.}
    \label{sec:definition}
    Phonemes are the smallest units of sound that can change a word's meaning, such as /p/ and /b/ in English \textit{pat} versus \textit{bat}. Every language uses a particular set of phonemes, known as its \emph{phoneme inventory} \citep{ladefoged1996sounds}. Some phonemes are shared across languages (e.g., /t/ occurs in both English and French), while others occur in only one of the two (e.g., the \emph{th} sound /\textipa{T}/ of English \emph{think} is absent from French).

\begin{figure}[t]
        \centering
        \includegraphics[width=\linewidth]{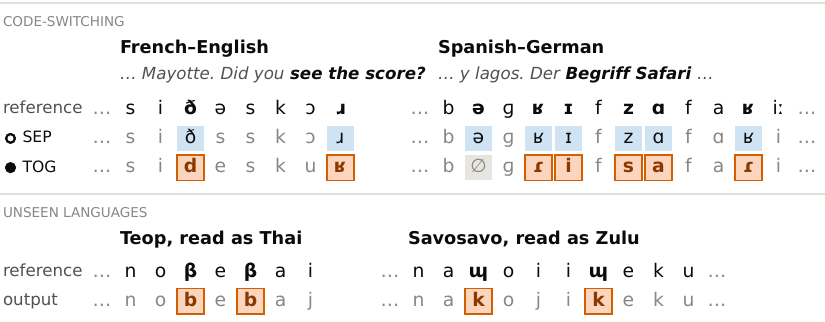}
        \caption{\textbf{Qualitative examples of interference.} \pxeus{} transcriptions aligned to the reference, with unshared phonemes in bold. Output at unshared phonemes is shaded blue when the phoneme is kept, orange with a frame and bold type when it is replaced by the other language's phoneme, and grey for any other error ($\varnothing$ marks a deletion). \textsc{sep} transcribes the second span without the other language in its context, and \textsc{tog} transcribes it with the other language in its context.
        In the French--English example, English \emph{``the score''} takes French [\textipa{d}] and \uvular{}.
        The unseen language rows (Section~\ref{sec:lowres}) have no \textsc{sep}/\textsc{tog} conditions, so we list the model-assigned language instead.
        }
        \label{fig:examples}
          \vspace{-1em}
    \end{figure}

We define \emph{phonological interference} as the loss of phonemes that one language has but another language lacks, in favor of that other language's phonemes; for example, English \rhotic{} may become French \uvular{} (more in Table~\ref{tab:unshared}, Appendix~\ref{app:exp_details}).
In code-switching, the other language is the second language of the utterance. For unseen languages, it is the training language the model assigns to the speech.
Bilingual speakers show interference around a code switch as small, gradient shifts toward the other language~\citep{antoniou2011inter, piccinini2015voice} (Appendix~\ref{app:interference}), while the models we study replace whole phonemes (Figure~\ref{fig:examples}).
We hypothesize that models estimate the language from context and impose its phoneme inventory, and test this in Section~\ref{sec:mechanism}.

\paragraph{Measurement.}
\label{sec:measurement}
We define the \emph{unshared phonemes} of a language pair $(A, B)$ as the phonemes of $A$ that $B$ lacks, and vice versa (Appendix~\ref{app:exp_details}). We measure how often the model preserves them (\emph{unshared phoneme recall}) and compare this against \emph{shared phoneme recall}, over the phonemes both languages have. If $B$ interferes with $A$, unshared recall should drop more than shared recall, whereas a general accuracy loss would lower both alike. An unshared phoneme of $A$ counts as lost only if it is deleted or replaced by a phoneme of $B$, since substitutions by any other phoneme cannot be attributed to interference. A shared phoneme counts as recalled only on an exact match.
    
    In code-switching, we focus on utterances consisting of one span in each of two languages. We process each span alone (\textsc{sep}) and both together (\textsc{tog}), and report the \textbf{interference} $\Delta = r_{\min}^{\textsc{sep}} - r_{\min}^{\textsc{tog}}$: the drop in unshared phoneme recall of the worse span, caused by the other language's presence. For unseen languages, we train a linear probe (Section~\ref{sec:estimate}) to identify which training language the model assigns to the recording (e.g., Teop is read as Thai in Figure~\ref{fig:examples}). Unshared phonemes are then the phonemes of the unseen language that this assigned training language lacks.

    \subsection{Models}
    \label{sec:models}
    
    \paragraph{Phone recognition.} We study two multilingual phone recognition models: POWSM~\citep{li2026powsm} and PhoneticXeus~\citep{pxeus26}. These are both encoder-based models with bidirectional attention over audio token embeddings; POWSM uses an E-Branchformer audio encoder with an autoregressive phoneme decoder, while PhoneticXeus uses the Xeus self-supervised audio encoder and non-autoregressive (CTC) phoneme prediction. POWSM is trained to predict a language tag as its first output token, and supports a \texttt{<unk>} language tag during inference for unseen or unknown languages. PhoneticXeus receives no language information during training.
    
    Both recognizers are trained on IPAPack++~\citep{zhu2025zipa}, roughly 17k hours of monolingual speech with phonemic transcriptions. The seven languages of our code-switched test sets are among its training languages, and none of the test recordings. Each span alone is therefore in distribution, and the code-switched recordings are new only in combining two seen languages. The languages of our third test set, DoReCo, are absent from it (Section~\ref{sec:data}).
    
    \paragraph{Text-to-Speech (TTS).} We use CosyVoice3 \citep{du2025cosyvoice}, an LLM-based TTS model built on a Qwen2~\citep{yang2024qwen2technicalreport} language model, adapted to generate speech tokens autoregressively followed by a flow matching module. We fine-tune the LLM part to accept IPA input on 30k monolingual IPAPack++~\citep{zhu2025zipa} utterances across 16 languages via LoRA, with no language tags or language embeddings, unlike multilingual TTS models that condition on the language explicitly~\citep{lou2025generalized}. The model receives only IPA and a fixed reference speaker audio. Each input phoneme is a token, embedded with a learned linear projection applied to its PanPhon articulatory features~\citep{mortensen2016panphon}.

    \label{sec:estimate}
    
    \paragraph{Language probe.} Several analyses need the language a model assigns to its input: which training language an unseen language is read as, and how confidently (Section~\ref{sec:lowres}), and how a second language in the context changes the estimate for a span (Section~\ref{sec:mechanism}). We train a 16-class logistic regression probe on the intermediate representations of monolingual IPAPack++ train split recordings, taken from each recognizer's audio encoder and from the TTS model's language model at its speech token positions. The probe gives each frame a score for each language, and we average these scores over the frames of a span or recording (Appendix~\ref{app:probe}; training details in Appendix~\ref{app:exp_details}).
    
    \subsection{Evaluation}
    \label{sec:evaluation}
    \label{sec:data}
    
    We evaluate on two code-switched datasets (specifically containing utterances that switch once between languages), and one dataset of monolingual speech in out-of-distribution languages. For datasets not already containing IPA transcripts, we obtain these by processing orthographic text transcripts to IPA with eSpeak NG~\citep{espeakng}.
    
    \paragraph{SwitchLingua.} We use 288 recordings of bilingual speakers reading code-switched sentences from SwitchLingua~\citep{xie2026switchlingua}, 48 per pairing of English with German, Spanish, French, Italian, Hindi or Korean. This dataset captures natural bilingual speech, but does not control for span length or language balance (e.g., the second span is usually shorter and in English). 
    
        \paragraph{Spliced FLEURS.} To complement SwitchLingua with a controlled set, we concatenate pairs of monolingual clips from the FLEURS test split~\citep{conneau2023fleurs}. We use the languages of SwitchLingua except French, which has no FLEURS recordings in the probe's training data, and create 10 samples for each ordered pair of six languages (300 concatenated samples total, about 8\,s each). Every language appears equally in both positions and duration.  
    
    \paragraph{DoReCo.} For unseen languages we use DoReCo~\citep{paschen2020building}, a collection of documentary recordings of low-resource languages with time aligned phonemic transcriptions. We use the version from the PRiSM benchmark~\citep{bharadwaj2026prism} covering 43 unseen languages.

We apply the interference metric (Section~\ref{sec:measurement}) to both code-switched datasets. On SwitchLingua, \textsc{sep} processes each span alone. On Spliced FLEURS, \textsc{sep} replaces the other-language half with a same-language half, so \textsc{sep} and \textsc{tog} differ only in the other half's language, with length and speaker change held constant. Processing each half alone gives similar results (Appendix~\ref{app:threecond}).
    
        To evaluate the code-switching datasets on phone recognition, we compare the model's IPA output directly to the reference. For TTS, we transcribe the synthesized audio back to IPA using PhoneticXeus. To prevent \pxeus{}'s own interference from contaminating the measurement, we split the audio at the language boundary (with forced alignment) and transcribe each language separately.

    \begin{figure}[t]
        \centering
        \includegraphics[width=\linewidth]{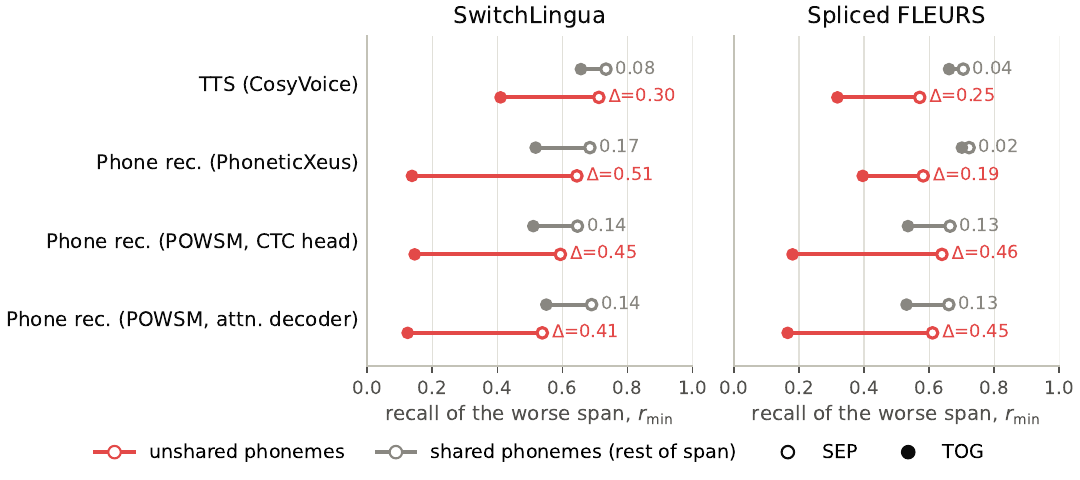}
        \caption{\textbf{Phonological interference on code-switched input.} Each bar runs from recall with each span processed separately (\textsc{sep}, hollow) to recall with both languages together (\textsc{tog}, filled); its length is the interference $\Delta$. Red: unshared phonemes. Grey: shared phonemes. Unshared phoneme recall drops much more than shared phoneme recall in every model and dataset (Table~\ref{tab:bleeding-detail}).}
        \label{fig:bleed}
          \vspace{-1em}
    \end{figure}

    \subsection{Results: Code-Switched Input}
    \label{sec:codeswitch-results}
    
Figure~\ref{fig:bleed} shows that interference is large and concentrated on unshared phonemes. Processing both languages together costs every model 32\% to 79\% of its unshared phoneme recall; under exact-match scoring, this drop is 1.9 to 8.2 times the drop in shared phonemes (Table~\ref{tab:bleeding-detail}). Most lost unshared phonemes are replaced by phonemes of the other language rather than by unrelated phonemes or deletions (Figure~\ref{fig:examples}): on SwitchLingua, 65\% of exact-match losses in TTS, 70\% in \pxeus{} and 78\% in \powsm{}'s CTC head. Some replacements could be merely notational, a similar sound written with a different symbol. However, in articulatory features, they are 3 to 4 times farther from the original phoneme than the other language's closest phoneme is (Appendix~\ref{app:subdistance}).

\paragraph{The same phoneme is lost more next to a language that does not have it.} Whether a phoneme is unshared depends on the language pair; English /\ipa{I}/, for example, is unshared next to Spanish and shared next to German. If the other language causes the loss, /\ipa{I}/ should lose more recall next to Spanish than next to German. We test this for every phoneme that is unshared with some languages and shared with others, with a logistic regression that holds the phoneme and the language pair fixed and compares the drop next to languages that lack the phoneme with the drop next to languages that have it (Table~\ref{tab:samephoneme}). On Spliced FLEURS, when the other language lacks the phoneme, the odds that the model recalls it fall by an extra factor of 2.2 in \powsm{} and 1.3 in \pxeus{}, compared with a language that has it. When we steer monolingual speech toward the other language (Section~\ref{sec:mechanism}), the extra factor is 9.8 and 11.5. For the TTS model, this Unshared-phoneme effect is not significant on Spliced FLEURS (CI includes 0) and is significant only in the steering experiment.

    \subsection{Results: Unseen Languages}
    \label{sec:lowres}
    
\begin{figure}[t]
        \centering
        \includegraphics[width=\linewidth]{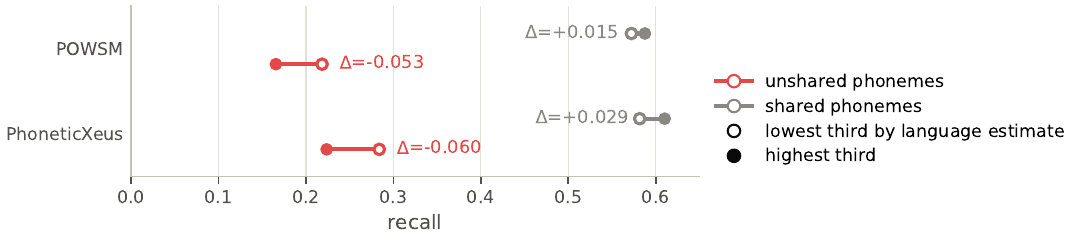}
        \caption{\textbf{Phonological interference in unseen languages.} 
        Recall on 43 DoReCo languages, split by the model's language probe confidence: lowest third (hollow) vs.\ highest third (filled). Unshared phoneme recall (red) drops the more the model is confident in the estimated language, while shared phonemes (grey) remain stable, isolating interference from general difficulty of unseen languages.}
        \label{fig:lowres}
        \vspace{-0.8em}
    \end{figure}

    Unseen language recordings contain no second language, so the \textsc{sep}/\textsc{tog} comparison does not apply. Instead, we compare recordings by how strongly the model commits to one of its training languages. For each recording, we take the probe's score (Section~\ref{sec:estimate}) for the training language assigned to that language (Section~\ref{sec:measurement}). Within each language, we rank the recordings by this score and compare the third with the lowest scores to the third with the highest. If interference comes from the language estimate, the top third should lose more of the phonemes that the assigned language lacks, and shared phonemes should be unaffected. Figure~\ref{fig:lowres} shows this pattern. From the lowest to the highest third, unshared phoneme recall falls by 0.053 (\powsm) and 0.060 (\pxeus{}), while shared phoneme recall rises slightly (+0.015 and +0.029), which rules out a general drop in audio quality. The lost phonemes are mostly replaced by phonemes of the assigned language (Appendix~\ref{app:lowres}). These results match the prediction that the model's language estimate drives interference.

    We evaluate only the phone recognizers here. The TTS output would have to be scored by transcribing it with \pxeus{}, which itself shows interference on unseen languages even without code-switching. Splitting at the language boundary, which protected the code-switched evaluation, does not help on monolingual input.

    \section{What Causes Phonological Interference?}
    \label{sec:mechanism}

    \begin{figure}[t]
        \centering
        \begin{subfigure}[t]{0.60\linewidth}
            \centering
            \includegraphics[width=\linewidth]{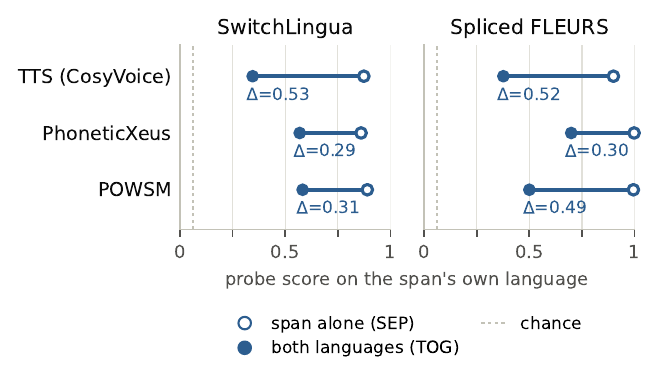}
            \caption{Language estimate on code-switched input}
            \label{fig:probedelta}
        \end{subfigure}\hfill
        \begin{subfigure}[t]{0.38\linewidth}
            \centering
            \includegraphics[width=\linewidth]{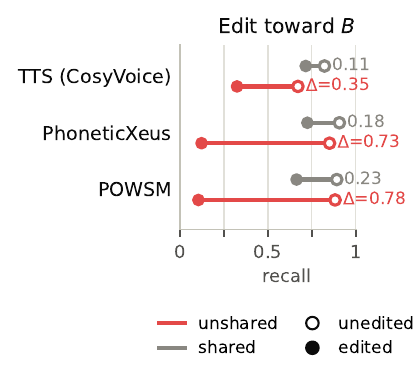}
            \caption{Editing the language subspace}
            \label{fig:steer}
        \end{subfigure}
        \caption{\textbf{The language estimate drops on code-switched input, and editing it reproduces interference.} (a) Probe score on each span's own language, processed separately (\textsc{sep}, hollow) vs.\ together (\textsc{tog}, filled), with the conditions of Section~\ref{sec:evaluation}. (b) Recall on monolingual utterances before (hollow) and after (filled) the language subspace is set to the mean of another language $B$. Red: unshared phonemes. Grey: shared phonemes. $\Delta$ is the drop in each panel.}
        \vspace{-1.2em}
    \end{figure}
    
    On code-switched input, the probe score (Section~\ref{sec:estimate}) for a span's own language drops substantially when the other language is present (Figure~\ref{fig:probedelta}), even though the audio of each span is unchanged. Motivated by this, we define the \emph{language subspace} directly from the language labels of the training data. At each block we take the centroid of each of the 16 probe languages, the mean hidden representation over that language's training frames. 
    The 16 centered centroids span 15 directions. We whiten them by the within-language covariance $\Sigma_w$, so that the subspace separates languages relative to the variation within each language, and take the top-$k$ singular directions $V \in \mathbb{R}^{D \times k}$ of the whitened centroids with $k = 15$, the full span, giving $\mathrm{proj}_{\mathrm{lang}}(\mathbf{h}) = \Sigma_w^{1/2} V V^\top \Sigma_w^{-1/2}\,\mathbf{h}$. This 15-dimensional subspace, under 2\% of each model's hidden dimensions, is the low dimensional language subspace we refer to throughout (Appendix~\ref{app:subspace}).
    
    To test whether this subspace causally drives interference, we manipulate it directly. Given a held-out monolingual utterance in language $A$, we edit it toward a target language $B$. Each block produces a representation $\mathbf{h}$ at every audio frame. At each frame and each block from a start block to the last, we replace the projection of $\mathbf{h}$ onto the language subspace with that of $\mathbf{c}_B$, the centroid representation of language $B$:
    \begin{equation}
        \mathbf{h} \leftarrow \mathbf{h} 
        - \mathrm{proj}_{\mathrm{lang}}(\mathbf{h}) 
        + \mathrm{proj}_{\mathrm{lang}}(\mathbf{c}_B),
        \label{eq:edit}
    \end{equation}
    leaving the rest of $\mathbf{h}$, $\mathbf{h} - \mathrm{proj}_{\mathrm{lang}}(\mathbf{h})$, unchanged. We apply the edit at blocks 6 to 8 of \powsm{}, 14 to 18 of \pxeus{} and 14 to 22 of the TTS model, and each later block computes from the edited output. We test on monolingual IPAPack++ utterances in all 16 probe languages, from held-out speakers; the TTS model receives the corresponding IPA input.

Figure~\ref{fig:steer} shows that this edit reproduces interference. The phone recognizers lose most of their unshared phonemes (86--88\% of their unedited recall) and the TTS model about half, while shared phonemes drop only a quarter to a third as much (Table~\ref{tab:steering-detail}, Appendix~\ref{app:subspace}). The effect lies almost entirely along the direction between the two languages: an edit along $A \to B$ alone produces 97\% of the unshared phoneme recall drop of the full 15-dimensional edit.
    
We ablate the editing effect with three controls. First, a random subspace of the same rank, shifted by the same amount, leaves recall almost unchanged: unshared phoneme $\Delta$ is $0.03$--$0.06$ across the three models, against $0.35$--$0.78$ for the language subspace. Second, erasing language information, by setting all frames to the mean of all 16 centroids, produces about half of the targeted edit's loss in the phone recognizers and 66\% of it in the TTS model ($\Delta$ $0.23$--$0.42$), without steering to any particular language. Third, which phonemes appear depends on the target: the edit toward $B$ adds $7.3$ (POWSM), $4.6$ (PhoneticXeus) and $0.9$ (TTS) phonemes per 100 that only $B$ uses, while the same edit toward a third language $C$ adds essentially none of $B$'s, yet removes $A$'s phonemes that $C$ lacks just as much ($\Delta$ $0.35$--$0.78$). Table~\ref{tab:steering-detail} (Appendix~\ref{app:subspace}) gives the full numbers with intervals.
    
\paragraph{The mechanism extends beyond phoneme-level models.} Automatic speech recognition (ASR) models such as Whisper~\citep{radford2022robustspeechrecognitionlargescale} transcribe speech to text, and perform worse on code-switched speech due to language confusion~\citep{liu2026code}. To test whether a similar language subspace can explain language confusion in standard ASR, we train the probe of Section~\ref{sec:estimate} on Whisper's encoder. We find that the other language lowers its score for the span's own language by $0.28$ on SwitchLingua, as in the phone recognizers, and by $0.11$ on Spliced FLEURS. We then edit the encoder's language subspace toward $B$ on monolingual speech in language $A$. This raises the probability of $B$'s language token (which Whisper predicts before transcribing) by $0.20$ and moves its transcript toward $B$, while editing toward a third language or through a random subspace does neither.
The same language subspace thus controls Whisper's choice of language (Appendix~\ref{app:whisper}).

    \section{Windowed Language Estimation Mitigates Interference}
    \label{sec:repair}
    
    To keep each language's phonemes in code-switched speech, the model must estimate the language locally, not from the whole utterance. We introduce \emph{windowed language estimation} (WLE), which replaces the language subspace coordinates at each frame with those computed from a local window and leaves the remaining coordinates unchanged, targeting interference in code-switching.
    
    We measure interference as in Section~\ref{sec:evaluation}, applying WLE to the whole recording (\textsc{tog}) and comparing it against each span decoded alone without any edit (\textsc{sep}), so that $\Delta$ also reflects any cost of the edit to the span itself. We report SwitchLingua in Table~\ref{tab:repair} and Spliced FLEURS in Table~\ref{tab:oracle} (Appendix~\ref{app:oracle}). We compare WLE with the same edit through a random subspace of equal rank and with a \emph{sliding window} that replaces every coordinate with the windowed one. For the recognizers, we measure the cost of WLE on monolingual speech as the change in two error rates on PRiSM~\citep{bharadwaj2026prism} (Table~\ref{tab:cost}). The phone error rate (PER) counts every substitution, deletion and insertion as one error, while the phone feature error rate (PFER), which PRiSM reports, weights each error by its articulatory feature distance, so a near miss costs less than a distant substitution. For the TTS model, we measure the cost as the change in recall of shared phonemes on the code-switched sentences, relative to generating each span alone (Appendix~\ref{app:repair}).
    
    \paragraph{Phone recognizers.} We encode the audio twice: once as a whole and once in 2\,s sliding windows with a 1\,s hop, each frame taken from the window in which it lies furthest from an edge. At each frame and at each block that Equation~\ref{eq:edit} edits, we replace the language subspace projection with that of the windowed encoding:
    \begin{equation}
        \mathbf{h} \leftarrow \mathbf{h} 
        - \mathrm{proj}_{\mathrm{lang}}(\mathbf{h}) 
        + \mathrm{proj}_{\mathrm{lang}}(\mathbf{h}_{\mathrm{win}}).
        \label{eq:wle}
    \end{equation}
    WLE reduces interference at window widths from 1 to 4\,s; Appendix~\ref{app:width} gives the full ablation. The windowed language estimate is computed from the model's own representations, so WLE does not require any external language identification system.
    
\paragraph{TTS model.} We apply the same edit (Equation~\ref{eq:wle}) to the TTS model, with a language subspace built the same way from its own hidden representations, at rank 15 and blocks 14 to 22 of its 24 (Appendix~\ref{app:subspace}). The difference is where $\mathbf{h}_{\mathrm{win}}$ comes from. The recognizers window their input audio; the TTS model's input is a phoneme sequence, so we window that instead. We generate the sentence in chunks of about 12 phonemes. For each chunk, we take the roughly 24 input phonemes centered on it, generate from this short sequence alone, with no history and no edit, and take the mean of that run's hidden states at each edited block as $\mathbf{h}_{\mathrm{win}}$. We then generate the chunk with the full phoneme input and all previously generated speech tokens as context, applying the edit at each of its positions (the first chunk is left unedited). We call this the \emph{fixed window}: like the recognizers' window, it spans about 2\,s of speech and needs no language label. We also test a \emph{span window}, in which the chunks are the two language spans themselves, split at the switch point in the input text; this uses the boundary but not the identity of either language. Neither generating in chunks without the edit nor applying the edit through a random subspace reduces interference.
    \begin{table}[t]
    \centering
    \small
    \begin{tabular}{llcc}
    \toprule
    Model & Condition & $r_{\min}^{\textsc{tog}} \uparrow$ & $\Delta$ (interference) $\downarrow$ \\
    \midrule
    \multirow{4}{*}{TTS}
      & No intervention & 0.410 & 0.302 \\
      & WLE, fixed window & 0.513 & 0.199 \\
      & \textbf{WLE}, span window & \textbf{0.619} & \textbf{0.093} \\
    \cmidrule(l){2-4}
      & \textcolor{gray}{\emph{Oracle}} & \textcolor{gray}{0.606} & \textcolor{gray}{0.106} \\
    \midrule
    \multirow{3}{*}{\pxeus{}}
      & No intervention & 0.138 & 0.507 \\
      & \textbf{WLE} & \textbf{0.430} & \textbf{0.214} \\
    \cmidrule(l){2-4}
      & \textcolor{gray}{\emph{Oracle}} & \textcolor{gray}{0.536} & \textcolor{gray}{0.108} \\
    \midrule
    \multirow{3}{*}{\powsm{}}
      & No intervention & 0.146 & 0.448 \\
      & \textbf{WLE} & \textbf{0.306} & \textbf{0.288} \\
    \cmidrule(l){2-4}
      & \textcolor{gray}{\emph{Oracle}} & \textcolor{gray}{0.472} & \textcolor{gray}{0.121} \\
    \bottomrule
    \end{tabular}
    \caption{\textbf{Effect of WLE on interference.} Best result in \textbf{bold}. WLE removes 69\% of the interference in TTS with the span window and 34\% with the fixed window, 58\% in \pxeus{}, and 36\% in \powsm{}. The \emph{Oracle} rows (gray) use the true language of every frame or span and serve as a reference (Appendix~\ref{app:oracle}). \powsm{} rows use its CTC head. $r_{\min}^{\textsc{tog}}$ and $\Delta$ as in Section~\ref{sec:evaluation}.}
    \label{tab:repair}
    \end{table}
    
    \begin{table}[t]
    \centering
    \small
    \begin{tabular}{lllcccc}
    \toprule
     & Model & Benchmark & \multicolumn{2}{c}{$\Delta$PER (pp) $\downarrow$} & \multicolumn{2}{c}{$\Delta$PFER (pp) $\downarrow$} \\
    \cmidrule(lr){4-5} \cmidrule(lr){6-7}
     & & & Sliding window & WLE & Sliding window & WLE \\
    \midrule
    \multirow{6}{*}{\rotatebox{90}{Seen}}
    & \multirow{3}{*}{\powsm{}}
    & TIMIT & $+0.96$ & $\mathbf{+0.07}$ & $+0.47$ & $\mathbf{+0.01}$ \\
    & & L2-ARCTIC & $+1.55$ & $\mathbf{+0.10}$ & $+0.31$ & $\mathbf{-0.00}$ \\
    & & GMU accent & $+2.14$ & $\mathbf{-0.32}$ & $+0.73$ & $\mathbf{-0.07}$ \\
    \cmidrule(lr){2-7}
    & \multirow{3}{*}{\pxeus{}}
    & TIMIT & $+1.33$ & $\mathbf{+0.08}$ & $+0.49$ & $\mathbf{+0.03}$ \\
    & & L2-ARCTIC & $+3.22$ & $\mathbf{+0.14}$ & $+0.50$ & $\mathbf{+0.03}$ \\
    & & GMU accent & $+2.71$ & $\mathbf{-0.09}$ & $+1.01$ & $\mathbf{-0.04}$ \\
    \midrule
    \multirow{2}{*}{\rotatebox{90}{\small Unsn.}}
    & \powsm{} & DoReCo & $+1.97$ & $\mathbf{-0.09}$ & $\mathbf{-0.45}$ & $+0.05$ \\
    \cmidrule(lr){2-7}
    & \pxeus{} & DoReCo & $+2.08$ & $\mathbf{-0.22}$ & $+1.06$ & $\mathbf{+0.15}$ \\
    \bottomrule
    \end{tabular}
    \caption{\textbf{Performance in monolingual speech.} Change in phone error rate ($\Delta$PER) and phone feature error rate ($\Delta$PFER), in percentage points, on monolingual benchmarks relative to unmodified decoding. Negative = improvement. The lower (better) value in each pair is in \textbf{bold}. The sliding window degrades PER and PFER much more than WLE, except for \powsm{}'s PFER on DoReCo.}
    \label{tab:cost}
    \vspace{-1.1em}
    \end{table}
    
\paragraph{Results.} WLE substantially reduces interference in all three models (Table~\ref{tab:repair}), and in the recognizers it stays within $\pm$0.35\,pp of baseline monolingual PER and $\pm$0.15\,pp of PFER (Table~\ref{tab:cost}). A naive sliding window over the full input, a less surgical approach to avoiding interference, also reduces interference but degrades monolingual PER by 1--3\,pp and PFER by up to 1.1\,pp. By targeting only the language subspace, WLE shows a favorable tradeoff, reducing interference at almost no cost to monolingual performance. The one exception is \powsm{} on DoReCo, where the sliding window lowers PFER by 0.45\,pp while it raises PER by 2.0\,pp. A random subspace of equal rank produces no meaningful change. In the TTS model the span window costs almost nothing on the shared phonemes; the fixed window costs more, since near the switch it contains phonemes of both languages (Appendix~\ref{app:repair}). Table~\ref{tab:wle-examples} shows sentences in which WLE undoes the replacement of unshared phonemes in all three models, such as English \rhotic{} becoming French \uvular{}.

    As a reference, the \emph{Oracle} rows of Table~\ref{tab:repair} apply the same edit with the true language of each frame or span in place of the windowed estimate. In the recognizers the oracle removes more interference than WLE, so a better local estimate could improve the repair further; in the TTS model the span window already matches it. The oracle requires the language of every frame as input, so it is impractical and cannot be used on unseen languages (Appendix~\ref{app:oracle}).

    \section{Related Work}
    \label{sec:rw}

    \paragraph{Language confusion in text LLMs.}
Multilingual language models encode language identity in a low dimensional subspace~\citep{chang2022geometry, xie2022discovering, dumas2025separating}, and LLMs generate text in the wrong language when this identity overrides the prompt~\citep{marchisio2024understanding, choo2026tlpo}. On code-switched input, their hidden states are anchored to one of the two languages~\citep{park2026code}. Existing fixes steer hidden states toward a target language given as input~\citep{yunfan2025mitigating, sterz2025recover}, or toward a source-language anchor extracted from the input~\citep{park2026code}. In speech, a single frame or IPA symbol carries little language identity, so the model must estimate the language from context. We show that speech models have a similar subspace, that one language interferes with the other through it, and that it also drives language confusion in ASR. WLE rewrites only this subspace with the model's own estimate from a short window, with no language label, so each span keeps its own language.
    
    \paragraph{Code-switching in speech models.}
    Code-switched speech degrades both recognition and synthesis~\citep{liu2026code, lee2025unicom, Kline2025ThePO}. \citet{liu2026code} identify language confusion as a key challenge in code-switched ASR, but measure it only through word level metrics, as do broad coverage code-switched benchmarks~\citep{yan2025cs}. \citet{stafford2026neural} document structured phoneme confusions without identifying their cause and \citet{lee2025unicom} propose data augmentation for recognition and translation.
    These works improve performance through data or model design. We measure phoneme-level interference in recognition and synthesis, on unseen languages and on code-switched input, trace its cause, and reduce it at inference time.
    
 \paragraph{Interpretability of speech models.}
Self-supervised speech models encode phonetic and phonemic information~\citep{pasad2021layer, choi2024self}, including a low dimensional subspace in HuBERT for a phonological distinction~\citep{martin2023probing}, and show some human-like native-language perceptual biases~\citep{millet2022self, rodriguez2024self}. These studies characterize what information is present without intervening on it. Like work on text LLMs that identifies and manipulates linear subspaces~\citep{marks2023geometry}, we show that editing a language subspace causally changes the output. WLE has the form of an interchange intervention, which replaces a subspace with its value from another run. Distributed alignment search~\citep{geiger2024finding} learns the subspace by optimization and copies its value from a different input to test a causal hypothesis. WLE instead uses the subspace spanned by the language centroids, which needs no optimization, and copies its value from a short window of the same input to repair the model at inference time.

        \section{Conclusion}
    
        We identified phonological interference in phoneme-level multilingual speech models: on code-switched input or languages absent from training, they impose the phonology of the language they estimate, replacing the phonemes it lacks with its own. We traced this to a low dimensional language subspace whose manipulation reproduces the effect and which also sets the output language of Whisper, linking interference to language confusion in ordinary speech recognition. Windowed language estimation substantially reduces interference on code-switched input without a language label, and leaves the phone recognizers' monolingual performance unchanged.
        
        \paragraph{Limitations and future work.}
        Our code-switched experiments use utterances that switch once between two languages, and extending to broader code-switching patterns remains future work. Our evidence on unseen languages compares recordings without intervening on the model. 
        Disentangling the contributions of model architecture, training data, and labeling conventions to the interference phenomenon is an important direction for future investigation.

        \paragraph{Acknowledgment.}
        We thank Odelia Schwartz, Nathaniel R. Robinson, Finn Ellingwood, Nimrod Shabtay, Natanel Nissan, Yakov Kolani, Kwanghee Choi and Michael Toker for the helpful discussions and feedback.

        \bibliography{iclr2027_conference}
        \bibliographystyle{iclr2027_conference}
            
        \newpage
        \appendix
        
        \section{Additional Experimental Details}
        \label{app:exp_details}
        \paragraph{Phone Recognizers.} We use the public checkpoints unchanged, on 16\,kHz mono audio. \powsm{} receives the \texttt{<unk>} language tag in every condition. We report its attention decoder (ESPnet defaults: beam 5, CTC weight 0) and greedy CTC decoding from the same encoder. \pxeus{} is decoded with greedy CTC. One encoder frame covers 40\,ms of audio in \powsm{} (25 frames per second) and 20\,ms in \pxeus{} (50 frames per second). One speech token of the TTS model covers 40\,ms (25 tokens per second).
    
        \paragraph{TTS training.} The base model is \texttt{Fun-CosyVoice3-0.5B-2512}, frozen. We train LoRA adapters (rank 32, $\alpha = 64$, dropout 0.05) on every attention and MLP projection of the LM, together with a linear projection of PanPhon features. The training set is 30,388 IPAPack++ utterances of 1 to 24\,s with at least five phonemes, about 2,000 per language (1,150 to 2,469), in English, French, Georgian, German, Greek, Hindi, Hungarian, Italian, Kinyarwanda, Korean, Polish, Spanish, Thai, Turkish, Vietnamese and Zulu. The targets are the tokens of CosyVoice3's speech tokenizer, and the loss is cross-entropy on the speech positions. We use a batch size of 8 for a single epoch with the official hyperparameters.
    
        \paragraph{TTS inference.} The LM input is the start token, the projected phonemes and the task token. We sample with CosyVoice3's default repetition aware sampling (top-$p$ 0.8, top-$k$ 25) and cap the output at 20 speech tokens (0.8 s) per input phoneme to avoid endless decodes, and leave the flow matching module and the vocoder unchanged. The reference speaker is the 3.5\,s prompt clip distributed with CosyVoice3. Each input is synthesized under four sampling seeds (0 to 3), and the recall of a span is averaged over seeds before the worse span is taken. For the \textsc{tog} condition the synthesized audio (24\,kHz) is force-aligned to its words and cut at the language boundary as described below.
    
        \paragraph{Language estimate.} The probe is a multinomial logistic regression on frames standardized with the training frames' mean and variance. It is fitted on 16,000 IPAPack++ utterances, 1,000 per language, drawn from Common Voice, an audiobook corpus (MLS, or LibriSpeech for English) and FLEURS where the language has them, and tested on 5,800 further utterances. The 2,389 training speakers and 1,112 test speakers are disjoint. Each utterance contributes at most 10 frames and each phoneme at most 2, located with the recognizer's CTC alignment, which gives up to 160,000 training frames. For the TTS a frame is one speech token position of the LM with the utterance's own speech tokens as input (teacher forcing). All audio is scaled to $-23$\,dBFS RMS. We read the probe at layer 8 of \powsm{}, layer 17 of \pxeus{} and block 7 of the TTS LM, the layers where the held-out probe score is highest (Appendix~\ref{app:probe}). 
    
        \paragraph{Reference transcriptions.} For SwitchLingua and FLEURS we transcribe the orthographic reference word by word with eSpeak NG 1.52 through \texttt{phonemizer} 3.4, each word in the voice of its own language (\texttt{en-us}, \texttt{de}, \texttt{es}, \texttt{fr-fr}, \texttt{hi}, \texttt{it}, \texttt{ko}). We remove stress marks and tie bars and apply a fixed symbol map per language that brings eSpeak's conventions to those of IPAPack++, for example German /r/ is written \ipa{K}, and length marks are dropped in English, Italian and Korean. 
    
        \paragraph{SwitchLingua sample.} For each language pair we shuffle the corpus recordings with a fixed seed and keep the first 48 whose sentence has 6 to 22 words and exactly one change of language between adjacent words, with at most three recordings from any one of the corpus's conversations. The 288 recordings come from 239 conversations.
    
        \paragraph{Cutting at the language boundary.} We align each recording to its words with \texttt{Qwen3-ForcedAligner-0.6B} and cut at the midpoint between the end of the last word of the first span and the start of the first word of the second. No audio is discarded or padded, and the same rule is applied to human recordings and synthesized audio.
    
        \paragraph{Spliced FLEURS.} Halves are taken from FLEURS test utterances with at least 4\,s of speech, with leading and trailing silence removed (samples below 2\% of the file's peak). The first half of a clip is the end of its utterance and the second half is the beginning of its utterance, so the join falls at natural utterance edges. The other side is cut at a word boundary that falls in a pause, located with the MMS forced aligner (\texttt{mms-300m-1130-forced-aligner}; Qwen3 for Korean), choosing the boundary that gives a half closest to 4\,s within 3 to 5\,s. The reference of a half is the phonemes of the words it contains. Each half receives a 5\,ms fade at both ends, and the two halves are joined with no gap or crossfade at 16\,kHz. We draw 10 clips for each of the 30 ordered language pairs with a fixed seed; the two halves of a clip are always different sentences. For every clip we also build the two same language splices used as \textsc{sep}, in which the other half is replaced by a half in the same language from a different speaker and sentence, giving 600 control clips.
    
        \paragraph{DoReCo.} PRiSM's DoReCo set has 18,299 clips in 44 languages. We exclude its French, which is one of the probe's 16 languages, and report on the remaining 43. The language assigned to a DoReCo language is the one with the highest probe score after averaging over frames, then over all of that language's clips. Clips are split into thirds by the rank of their probe score for the assigned language within their own language. A phoneme is unshared when the DoReCo references of the language contain it at least three times and the probe's training references of the assigned language contain it fewer than three times.
    
        \paragraph{Scoring.} The unshared phonemes of a language pair are those that occur at least three times in that language's reference spans and never in the other language's, counted over the pair's recordings in the dataset. Reference and output are aligned by edit distance with unit costs. A phoneme is recalled when the aligned output phoneme is identical after normalization (Appendix~\ref{app:unmarked}); a deletion is a loss. For unshared phonemes, a substitution by a phoneme absent from the other language's reference spans is also accepted, so the losses are deletions and substitutions by the other language's phonemes. The share of losses replaced by the other language's phonemes (Section~\ref{sec:bleed-results}) is the rise in such substitutions from \textsc{sep} to \textsc{tog} over the fall in exact matches. Recordings in which one span has no unshared phoneme are not scored, which leaves 266 of 288 SwitchLingua recordings (265 for \pxeus{}, which lacks one decode) and 282 of 300 spliced clips (274 for \powsm{}'s attention decoder, which returns 8 empty decodes). Intervals are 95\% percentile bootstrap intervals from 10,000 resamples of recordings.
    
        Table~\ref{tab:unshared} shows examples and their most common replacements, and Table~\ref{tab:unsharedfull} lists the resulting sets.

        \begin{table}[t]
        \centering
        \small
        \renewcommand{\arraystretch}{1.15}
        \begin{tabular}{@{}l c@{\hspace{0.3em}}c@{\hspace{0.3em}}c @{\hspace{1em}} l @{\hspace{2.5em}} l c@{\hspace{0.3em}}c@{\hspace{0.3em}}c @{\hspace{1em}} l@{}}
        \toprule
        Language pair & \multicolumn{3}{c}{Replacement} & Example & Language pair & \multicolumn{3}{c}{Replacement} & Example \\
        \midrule
        English--French & \ipa{\textturnr{}} & $\rightarrow$ & \ipa{K} & \textit{\textbf{r}eally} & German--Spanish & \ipa{I} & $\rightarrow$ & \ipa{i} & \textit{m\textbf{i}t} \\
        English--German & \ipa{w} & $\rightarrow$ & \ipa{v} & \textit{\textbf{w}as} & English--German & \ipa{D} & $\rightarrow$ & \ipa{d} & \textit{\textbf{th}e} \\
        English--Spanish & \ipa{A} & $\rightarrow$ & \ipa{o} & \textit{pr\textbf{o}bably} & English--Spanish & \ipa{z} & $\rightarrow$ & \ipa{s} & \textit{i\textbf{s}} \\
        German--Hindi & \ipa{K} & $\rightarrow$ & \ipa{R} & \textit{de\textbf{r}} & English--Italian & \ipa{\ae{}} & $\rightarrow$ & \ipa{a} & \textit{\textbf{a}nd} \\
        \bottomrule
        \end{tabular}
        \caption{\textbf{Examples of unshared phonemes} from language pairs in our experiments. Each phoneme at the left of an arrow occurs only in the first language of the pair; the arrow shows its most common replacement when both languages are processed together, with an example word whose bold letters spell the phoneme. Table~\ref{tab:unsharedfull} lists the full sets for every pair and dataset.
        }
        \label{tab:unshared}
        \end{table}
    
        \begin{table}[t]
        \centering
        \small
        \begin{tabular}{@{}l p{0.41\linewidth} p{0.41\linewidth}@{}}
        \toprule
        Pair (A--B) & Only in A & Only in B \\
        \midrule
        \multicolumn{3}{@{}l}{\textit{SwitchLingua}} \\
        English--French & \ipa{\*r}\textsubscript{54} \ipa{D}\textsubscript{33} \ipa{\textrhookschwa{}}\textsubscript{30} \ipa{\ae{}}\textsubscript{29} \ipa{2}\textsubscript{21} \ipa{h}\textsubscript{18} & \ipa{K}\textsubscript{141} \ipa{y}\textsubscript{45} \ipa{\oe{}}\textsubscript{11} \ipa{\o{}}\textsubscript{4} \\
        English--German & \ipa{\*r}\textsubscript{53} \ipa{w}\textsubscript{28} \ipa{\ae{}}\textsubscript{28} \ipa{D}\textsubscript{28} \ipa{\textrhookschwa{}}\textsubscript{18} \ipa{T}\textsubscript{8} \ipa{Z}\textsubscript{6} & \ipa{K}\textsubscript{124} \ipa{\c{c}}\textsubscript{44} \ipa{5}\textsubscript{35} \ipa{\o{}}\textsubscript{13} \ipa{x}\textsubscript{11} \ipa{y}\textsubscript{10} \ipa{\oe{}}\textsubscript{5} \ipa{Y}\textsubscript{5} \\
        English--Hindi & \ipa{f}\textsubscript{35} \ipa{D}\textsubscript{33} \ipa{\ae{}}\textsubscript{31} \ipa{\textrhookschwa{}}\textsubscript{27} \ipa{2}\textsubscript{24} \ipa{w}\textsubscript{21} \ipa{A}\textsubscript{17} \ipa{v}\textsubscript{15} \ipa{T}\textsubscript{7} & \ipa{R}\textsubscript{61} \ipa{V}\textsubscript{19} \ipa{k\super{h}}\textsubscript{18} \ipa{b\super{h}}\textsubscript{12} \ipa{t\super{h}}\textsubscript{11} \ipa{S\super{h}}\textsubscript{9} \ipa{p\super{h}}\textsubscript{8} \ipa{\:t}\textsubscript{7} \ipa{\:d}\textsubscript{6} \ipa{Z\super{h}}\textsubscript{3} \\
        English--Italian & \ipa{I}\textsubscript{105} \ipa{\*r}\textsubscript{54} \ipa{@}\textsubscript{38} \ipa{U}\textsubscript{30} \ipa{2}\textsubscript{27} \ipa{D}\textsubscript{18} \ipa{N}\textsubscript{15} \ipa{\ae{}}\textsubscript{11} \ipa{A}\textsubscript{11} \ipa{\textrhookschwa{}}\textsubscript{11} \ipa{T}\textsubscript{10} \ipa{h}\textsubscript{6} & \ipa{r}\textsubscript{173} \ipa{L}\textsubscript{5} \\
        English--Korean & \ipa{@}\textsubscript{63} \ipa{\*r}\textsubscript{62} \ipa{U}\textsubscript{38} \ipa{\textrhookschwa{}}\textsubscript{35} \ipa{f}\textsubscript{30} \ipa{D}\textsubscript{27} \ipa{\ae{}}\textsubscript{26} \ipa{v}\textsubscript{17} \ipa{O}\textsubscript{14} \ipa{A}\textsubscript{12} \ipa{Z}\textsubscript{5} \ipa{T}\textsubscript{3} & \ipa{W}\textsubscript{119} \ipa{R}\textsubscript{77} \ipa{\textltilde{}}\textsubscript{57} \ipa{C}\textsubscript{53} \ipa{\textctz{}}\textsubscript{47} \\
        English--Spanish & \ipa{I}\textsubscript{113} \ipa{\*r}\textsubscript{46} \ipa{U}\textsubscript{33} \ipa{z}\textsubscript{23} \ipa{e}\textsubscript{22} \ipa{D}\textsubscript{20} \ipa{A}\textsubscript{20} \ipa{O}\textsubscript{19} \ipa{2}\textsubscript{15} \ipa{v}\textsubscript{14} \ipa{\ae{}}\textsubscript{12} \ipa{\textrhookschwa{}}\textsubscript{11} \ipa{T}\textsubscript{8} \ipa{Z}\textsubscript{7} \ipa{h}\textsubscript{6} & \ipa{R}\textsubscript{138} \ipa{r}\textsubscript{16} \ipa{x}\textsubscript{16} \ipa{\textltailn{}}\textsubscript{5} \\
        \midrule
        \multicolumn{3}{@{}l}{\textit{Spliced FLEURS}} \\
        English--German & \ipa{\*r}\textsubscript{70} \ipa{D}\textsubscript{50} \ipa{\textrhookschwa{}}\textsubscript{42} \ipa{\ae{}}\textsubscript{41} \ipa{2}\textsubscript{30} \ipa{w}\textsubscript{15} \ipa{Z}\textsubscript{13} \ipa{T}\textsubscript{6} & \ipa{K}\textsubscript{107} \ipa{5}\textsubscript{38} \ipa{\c{c}}\textsubscript{33} \ipa{x}\textsubscript{14} \ipa{y}\textsubscript{13} \ipa{\o{}}\textsubscript{5} \ipa{\oe{}}\textsubscript{4} \\
        English--Hindi & \ipa{\*r}\textsubscript{68} \ipa{D}\textsubscript{50} \ipa{\ae{}}\textsubscript{36} \ipa{2}\textsubscript{36} \ipa{\textrhookschwa{}}\textsubscript{35} \ipa{f}\textsubscript{29} \ipa{v}\textsubscript{28} \ipa{w}\textsubscript{22} \ipa{N}\textsubscript{19} \ipa{A}\textsubscript{19} \ipa{T}\textsubscript{10} & \ipa{R}\textsubscript{96} \ipa{V}\textsubscript{38} \ipa{b\super{h}}\textsubscript{15} \ipa{\:t}\textsubscript{15} \ipa{t\super{h}}\textsubscript{8} \ipa{\:d}\textsubscript{7} \ipa{d\super{h}}\textsubscript{6} \ipa{\:n}\textsubscript{6} \ipa{k\super{h}}\textsubscript{5} \ipa{p\super{h}}\textsubscript{5} \ipa{S\super{h}}\textsubscript{5} \\
        English--Italian & \ipa{I}\textsubscript{187} \ipa{@}\textsubscript{106} \ipa{\*r}\textsubscript{88} \ipa{\ae{}}\textsubscript{47} \ipa{2}\textsubscript{47} \ipa{D}\textsubscript{44} \ipa{U}\textsubscript{37} \ipa{\textrhookschwa{}}\textsubscript{36} \ipa{A}\textsubscript{33} \ipa{h}\textsubscript{17} & \ipa{r}\textsubscript{111} \\
        English--Korean & \ipa{\*r}\textsubscript{106} \ipa{@}\textsubscript{89} \ipa{z}\textsubscript{51} \ipa{\textrhookschwa{}}\textsubscript{42} \ipa{\ae{}}\textsubscript{41} \ipa{O}\textsubscript{38} \ipa{v}\textsubscript{29} \ipa{A}\textsubscript{28} \ipa{Z}\textsubscript{8} \ipa{T}\textsubscript{4} & \ipa{W}\textsubscript{135} \ipa{R}\textsubscript{71} \ipa{\textltilde{}}\textsubscript{40} \ipa{C}\textsubscript{30} \ipa{\textctz{}}\textsubscript{23} \\
        English--Spanish & \ipa{I}\textsubscript{176} \ipa{@}\textsubscript{121} \ipa{\*r}\textsubscript{90} \ipa{z}\textsubscript{66} \ipa{\ae{}}\textsubscript{55} \ipa{D}\textsubscript{45} \ipa{2}\textsubscript{40} \ipa{U}\textsubscript{38} \ipa{\textrhookschwa{}}\textsubscript{36} \ipa{e}\textsubscript{26} \ipa{O}\textsubscript{26} \ipa{v}\textsubscript{25} \ipa{A}\textsubscript{19} \ipa{h}\textsubscript{18} \ipa{Z}\textsubscript{8} \ipa{T}\textsubscript{8} & \ipa{R}\textsubscript{125} \ipa{r}\textsubscript{22} \ipa{x}\textsubscript{18} \ipa{J}\textsubscript{5} \ipa{\textltailn{}}\textsubscript{3} \ipa{L}\textsubscript{3} \\
        German--Hindi & \ipa{K}\textsubscript{125} \ipa{v}\textsubscript{44} \ipa{A}\textsubscript{31} \ipa{\c{c}}\textsubscript{25} \ipa{5}\textsubscript{21} \ipa{x}\textsubscript{12} \ipa{y}\textsubscript{10} \ipa{\o{}}\textsubscript{10} \ipa{Y}\textsubscript{9} \ipa{\oe{}}\textsubscript{4} & \ipa{R}\textsubscript{90} \ipa{Z}\textsubscript{28} \ipa{V}\textsubscript{20} \ipa{d\super{h}}\textsubscript{13} \ipa{b\super{h}}\textsubscript{12} \ipa{t\super{h}}\textsubscript{12} \ipa{k\super{h}}\textsubscript{11} \ipa{\:d}\textsubscript{10} \ipa{\:t}\textsubscript{8} \ipa{\:n}\textsubscript{6} \ipa{\:s}\textsubscript{5} \ipa{S\super{h}}\textsubscript{3} \\
        German--Italian & \ipa{@}\textsubscript{176} \ipa{I}\textsubscript{128} \ipa{K}\textsubscript{122} \ipa{U}\textsubscript{68} \ipa{A}\textsubscript{37} \ipa{5}\textsubscript{35} \ipa{\c{c}}\textsubscript{31} \ipa{x}\textsubscript{16} \ipa{h}\textsubscript{15} \ipa{y}\textsubscript{11} \ipa{\oe{}}\textsubscript{9} \ipa{\o{}}\textsubscript{8} \ipa{Y}\textsubscript{8} & \ipa{r}\textsubscript{115} \ipa{w}\textsubscript{10} \ipa{L}\textsubscript{5} \\
        German--Korean & \ipa{@}\textsubscript{181} \ipa{K}\textsubscript{129} \ipa{z}\textsubscript{51} \ipa{U}\textsubscript{48} \ipa{O}\textsubscript{45} \ipa{f}\textsubscript{44} \ipa{5}\textsubscript{39} \ipa{v}\textsubscript{37} \ipa{\c{c}}\textsubscript{35} \ipa{A}\textsubscript{31} \ipa{x}\textsubscript{10} \ipa{y}\textsubscript{8} \ipa{\o{}}\textsubscript{7} \ipa{Y}\textsubscript{5} \ipa{\oe{}}\textsubscript{4} & \ipa{W}\textsubscript{117} \ipa{2}\textsubscript{90} \ipa{R}\textsubscript{63} \ipa{\textltilde{}}\textsubscript{43} \ipa{w}\textsubscript{32} \ipa{C}\textsubscript{29} \ipa{\textctz{}}\textsubscript{28} \\
        German--Spanish & \ipa{@}\textsubscript{148} \ipa{K}\textsubscript{121} \ipa{I}\textsubscript{113} \ipa{U}\textsubscript{54} \ipa{v}\textsubscript{40} \ipa{z}\textsubscript{40} \ipa{A}\textsubscript{37} \ipa{5}\textsubscript{30} \ipa{\c{c}}\textsubscript{29} \ipa{e}\textsubscript{27} \ipa{O}\textsubscript{24} \ipa{y}\textsubscript{17} \ipa{h}\textsubscript{13} \ipa{\o{}}\textsubscript{6} \ipa{Y}\textsubscript{5} & \ipa{R}\textsubscript{112} \ipa{r}\textsubscript{16} \ipa{w}\textsubscript{14} \ipa{J}\textsubscript{8} \ipa{\textltailn{}}\textsubscript{5} \ipa{L}\textsubscript{3} \\
        Hindi--Italian & \ipa{@}\textsubscript{170} \ipa{R}\textsubscript{119} \ipa{I}\textsubscript{67} \ipa{U}\textsubscript{25} \ipa{V}\textsubscript{23} \ipa{\:t}\textsubscript{17} \ipa{S\super{h}}\textsubscript{9} \ipa{d\super{h}}\textsubscript{6} \ipa{k\super{h}}\textsubscript{6} \ipa{\:d}\textsubscript{6} \ipa{b\super{h}}\textsubscript{5} \ipa{p\super{h}}\textsubscript{4} \ipa{t\super{h}}\textsubscript{4} \ipa{\:n}\textsubscript{4} \ipa{\:s}\textsubscript{3} & \ipa{r}\textsubscript{106} \ipa{v}\textsubscript{19} \ipa{w}\textsubscript{10} \ipa{L}\textsubscript{8} \ipa{N}\textsubscript{4} \\
        Hindi--Korean & \ipa{@}\textsubscript{187} \ipa{Z}\textsubscript{32} \ipa{V}\textsubscript{26} \ipa{U}\textsubscript{25} \ipa{\:d}\textsubscript{19} \ipa{O}\textsubscript{17} \ipa{b\super{h}}\textsubscript{12} \ipa{\:t}\textsubscript{12} \ipa{f}\textsubscript{10} \ipa{\:n}\textsubscript{10} \ipa{d\super{h}}\textsubscript{9} \ipa{t\super{h}}\textsubscript{8} \ipa{k\super{h}}\textsubscript{6} \ipa{z}\textsubscript{4} \ipa{g\super{h}}\textsubscript{3} \ipa{S\super{h}}\textsubscript{3} & \ipa{W}\textsubscript{131} \ipa{2}\textsubscript{126} \ipa{\textltilde{}}\textsubscript{52} \ipa{\textctz{}}\textsubscript{31} \ipa{C}\textsubscript{27} \\
        Hindi--Spanish & \ipa{@}\textsubscript{178} \ipa{e}\textsubscript{95} \ipa{I}\textsubscript{65} \ipa{h}\textsubscript{62} \ipa{Z}\textsubscript{23} \ipa{U}\textsubscript{22} \ipa{V}\textsubscript{20} \ipa{O}\textsubscript{17} \ipa{\:t}\textsubscript{15} \ipa{b\super{h}}\textsubscript{9} \ipa{d\super{h}}\textsubscript{8} \ipa{\:d}\textsubscript{8} \ipa{k\super{h}}\textsubscript{6} \ipa{t\super{h}}\textsubscript{6} \ipa{S\super{h}}\textsubscript{4} \ipa{z}\textsubscript{3} \ipa{\:n}\textsubscript{3} \ipa{\:s}\textsubscript{3} & \ipa{x}\textsubscript{16} \ipa{f}\textsubscript{13} \ipa{r}\textsubscript{13} \ipa{w}\textsubscript{12} \ipa{L}\textsubscript{4} \ipa{J}\textsubscript{3} \\
        Italian--Korean & \ipa{r}\textsubscript{100} \ipa{v}\textsubscript{23} \ipa{O}\textsubscript{20} \ipa{f}\textsubscript{15} \ipa{\textltailn{}}\textsubscript{7} \ipa{L}\textsubscript{5} \ipa{Z}\textsubscript{5} & \ipa{W}\textsubscript{119} \ipa{2}\textsubscript{96} \ipa{h}\textsubscript{86} \ipa{R}\textsubscript{71} \ipa{\textltilde{}}\textsubscript{45} \ipa{C}\textsubscript{38} \ipa{\textctz{}}\textsubscript{29} \\
        Italian--Spanish & \ipa{e}\textsubscript{134} \ipa{O}\textsubscript{19} \ipa{v}\textsubscript{13} \ipa{z}\textsubscript{11} \ipa{Z}\textsubscript{8} & \ipa{R}\textsubscript{106} \ipa{x}\textsubscript{23} \ipa{J}\textsubscript{5} \ipa{\textltailn{}}\textsubscript{4} \\
        Korean--Spanish & \ipa{W}\textsubscript{104} \ipa{2}\textsubscript{99} \ipa{h}\textsubscript{95} \ipa{\textltilde{}}\textsubscript{35} \ipa{\textctz{}}\textsubscript{32} \ipa{C}\textsubscript{31} \ipa{e}\textsubscript{27} \ipa{I}\textsubscript{3} & \ipa{x}\textsubscript{13} \ipa{r}\textsubscript{10} \ipa{L}\textsubscript{5} \ipa{J}\textsubscript{4} \\
        \bottomrule
        \end{tabular}
        \caption{\textbf{The full unshared phoneme sets} of every language pair in both code-switched datasets. Since selected phonemes depend on an occurrence threshold in observed data, the exact phonemes selected may differ between the datasets for the same pair of languages (e.g., comparing English--German in both datasets). Phonemes are ordered from most to least frequent, and the subscript is the number of occurrences in those reference spans.}
        \label{tab:unsharedfull}
        \end{table}
    
        \section{Detailed Interference Results}
        \label{app:unmarked}
        
        Table~\ref{tab:bleeding-detail} gives the full numbers behind Figure~\ref{fig:bleed}. Before comparing reference and output phonemes, we normalize diacritics, including length, nasalization, non-syllabic marks, and any diacritic appearing fewer than three times in the reference transcriptions.

    \begin{table}[t]
    \centering
    \small
    \setlength{\tabcolsep}{3pt}
    
    \textbf{(a) SwitchLingua} (265/266 scored)
    
    \vspace{0.3em}
    \begin{tabular}{l ccc cc c}
    \toprule
    & \multicolumn{3}{c}{\textbf{Unshared phonemes}} 
    & \multicolumn{2}{c}{\textbf{Shared phonemes}} 
    & \\
    \cmidrule(lr){2-4} \cmidrule(lr){5-6}
    Model 
    & $\Delta$ & $\Delta/r^{\textsc{sep}}$ & $\Delta_{\text{exact}}$ 
    & $\Delta$ & $\Delta/r^{\textsc{sep}}$ 
    & Ratio \\
    \midrule
    TTS (CosyVoice)          
      & .302 {\scriptsize[.265,\,.340]} & .424 & .224 {\scriptsize[.186,\,.262]} 
      & .077 {\scriptsize[.065,\,.090]} & .105 
      & 2.9$\times$ \\
    Phone rec.\ (PhoneticXeus) 
      & .507 {\scriptsize[.462,\,.551]} & .786 & .345 {\scriptsize[.304,\,.387]} 
      & .167 {\scriptsize[.145,\,.188]} & .244 
      & 2.1$\times$ \\
    Phone rec.\ (POWSM, CTC)  
      & .448 {\scriptsize[.404,\,.490]} & .754 & .365 {\scriptsize[.327,\,.402]} 
      & .136 {\scriptsize[.110,\,.161]} & .210 
      & 2.7$\times$ \\
    Phone rec.\ (POWSM, attn.) 
      & .414 {\scriptsize[.371,\,.456]} & .769 & .268 {\scriptsize[.233,\,.304]} 
      & .139 {\scriptsize[.116,\,.164]} & .202 
      & 1.9$\times$ \\
    \bottomrule
    \end{tabular}
    
    \vspace{1em}
    
    \textbf{(b) Spliced FLEURS} (274/282 scored)
    
    \vspace{0.3em}
    \begin{tabular}{l ccc cc c}
    \toprule
    & \multicolumn{3}{c}{\textbf{Unshared phonemes}} 
    & \multicolumn{2}{c}{\textbf{Shared phonemes}} 
    & \\
    \cmidrule(lr){2-4} \cmidrule(lr){5-6}
    Model 
    & $\Delta$ & $\Delta/r^{\textsc{sep}}$ & $\Delta_{\text{exact}}$ 
    & $\Delta$ & $\Delta/r^{\textsc{sep}}$ 
    & Ratio \\
    \midrule
    TTS (CosyVoice)          
      & .253 {\scriptsize[.222,\,.284]} & .443 & .198 {\scriptsize[.169,\,.228]} 
      & .044 {\scriptsize[.033,\,.055]} & .062 
      & 4.5$\times$ \\
    Phone rec.\ (PhoneticXeus) 
      & .186 {\scriptsize[.156,\,.218]} & .320 & .181 {\scriptsize[.150,\,.213]} 
      & .022 {\scriptsize[.009,\,.036]} & .030 
      & 8.2$\times$ \\
    Phone rec.\ (POWSM, CTC)  
      & .459 {\scriptsize[.425,\,.494]} & .718 & .414 {\scriptsize[.384,\,.445]} 
      & .129 {\scriptsize[.112,\,.148]} & .195 
      & 3.2$\times$ \\
    Phone rec.\ (POWSM, attn.) 
      & .445 {\scriptsize[.408,\,.482]} & .730 & .421 {\scriptsize[.388,\,.453]} 
      & .130 {\scriptsize[.105,\,.156]} & .197 
      & 3.2$\times$ \\
    \bottomrule
    \end{tabular}
    
    \caption{\textbf{Phonological interference on code-switched input} (full numbers behind Figure~3),
    with 95\% bootstrap intervals.
    $\Delta$: drop in worst span recall from \textsc{sep} to \textsc{tog}.
    $\Delta/r^{\textsc{sep}}_{\min}$: drop as a fraction of the \textsc{sep} level.
    $\Delta_{\text{exact}}$: the same drop under exact match, the rule used for shared phonemes; unshared
    $\Delta$ also accepts substitutions by phonemes the other language lacks.
    Ratio: unshared $\Delta_{\text{exact}}$ divided by shared $\Delta$.}
    \label{tab:bleeding-detail}
    \end{table}

        \paragraph{Choice of the \textsc{sep} baseline on Spliced FLEURS.}
        \label{app:threecond}
        On Spliced FLEURS each half is processed alone, next to a same language half, and next to the other language half. Table~\ref{tab:threecond} gives the worst span recall of unshared phonemes in each condition. A same language half changes recall by at most 0.012, and every interval includes 0, so $\Delta$ is nearly the same from either baseline.
    
        \begin{table}[t]
        \centering
        \small
        \setlength{\tabcolsep}{3pt}
        \begin{tabular}{l ccc cc}
        \toprule
        & \multicolumn{3}{c}{$r_{\min}$} & \multicolumn{2}{c}{$\Delta$ from} \\
        \cmidrule(lr){2-4} \cmidrule(lr){5-6}
        Model & Alone & Same lang. & Other lang. & Alone & Same lang. \\
        \midrule
        TTS (CosyVoice) & .570 & .571 & .318 & .252 {\scriptsize[.217,\,.287]} & .253 {\scriptsize[.222,\,.284]} \\
        Phone rec.\ (PhoneticXeus) & .591 & .582 & .396 & .196 {\scriptsize[.165,\,.228]} & .186 {\scriptsize[.156,\,.218]} \\
        Phone rec.\ (POWSM, CTC) & .636 & .639 & .180 & .456 {\scriptsize[.421,\,.491]} & .459 {\scriptsize[.425,\,.494]} \\
        Phone rec.\ (POWSM, attn.) & .598 & .610 & .165 & .433 {\scriptsize[.394,\,.472]} & .445 {\scriptsize[.408,\,.482]} \\
        \bottomrule
        \end{tabular}
        \caption{\textbf{Two \textsc{sep} baselines on Spliced FLEURS.} Worst span recall of unshared phonemes with each half alone, next to a same language half (\textsc{sep} in the main text) and next to the other language half (\textsc{tog}), and the drop to \textsc{tog} from each baseline, with 95\% bootstrap intervals.}
        \label{tab:threecond}
        \end{table}

\paragraph{The same phoneme next to different languages.}
\label{app:samephoneme}
Section~\ref{sec:codeswitch-results} reports that a phoneme loses more recall
next to a language that lacks it. An alternative explanation is that unshared
phonemes are simply harder to recognize or synthesize, and would lose more
recall next to any language. To separate the two, we use phonemes that are
unshared in some language pairs and shared in others. English /\ipa{I}/, for
example, is unshared next to Spanish, which lacks it, and shared next to
German, which has it. It is the same English sound in both pairs, so if it
loses more recall next to Spanish than next to German, the difference comes
from the other language and not from the phoneme itself.

Only phonemes that are unshared in at least one pair and shared in at least
one other contribute to this comparison. On Spliced FLEURS there are 56, about half of them vowels and taps (brief \emph{r}-like sounds), such as English
/\ipa{I}/ and /\ipa{@}/ and Hindi /\ipa{R}/. A phoneme that no other language
in the set has, such as English /\ipa{\*r}/, is unshared in every pair and
therefore cannot enter the comparison.

\textit{Data.} For every phoneme in the reference transcript, we record whether
the output reproduces it exactly, once before and once after the second
language is introduced. ``Before'' is a half spliced next to another half in
its own language, and ``after'' is the same half spliced next to the other
language.

\textit{Model.} Consider a phoneme $p$ in a half of language $A$, spliced next
to a half of language $B$. Let $u = 1$ if $B$ lacks the phoneme $p$ and $u = 0$ if $B$ has
it. Each reference phoneme is scored twice, and $t$ marks the condition:
$t = 0$ before the second language is introduced and $t = 1$ after. We fit one logistic regression
that predicts whether $p$ is recalled:
\[
  \operatorname{logit} \Pr(\text{recalled})
  =
  \underbrace{a_{p,A} + b_{A,B} + d\,u}_{\text{before}}
  -
  t\bigl(
    \underbrace{\alpha_{p,A} + \beta_{A,B} + \gamma\,u}_{\text{drop}}
  \bigr),
\]
where $\operatorname{logit}(x) = \log \frac{x}{1-x}$ is the log-odds. The first
group of terms gives the log-odds of recall before the second language is
introduced, and the second group gives how far they fall afterwards.

\textit{Quantity of interest.} $\gamma$ is how much further the log-odds of
recall fall when $B$ lacks $p$ than when $B$ has it. If the drop depended only
on the phoneme, and not on whether the other language has it, $\gamma$ would
be 0. We also report $e^{\gamma}$, which is easier to read: the odds of recall,
$\Pr / (1 - \Pr)$, fall by an extra factor of $e^{\gamma}$ next to a language
that lacks the phoneme.

\textit{Controls.} The remaining terms absorb other reasons recall might
differ. $a_{p,A}$ and $\alpha_{p,A}$ give each phoneme its own recall before
and its own drop, separately per language (English /\ipa{t}/ and German
/\ipa{t}/ get separate terms). $b_{A,B}$ and $\beta_{A,B}$ do the same for each
ordered language pair, absorbing pairs that are harder overall. $d$ lets
phonemes the other language lacks start from a different recall, so that a
lower starting point is not mistaken for a larger drop. Because
$\alpha_{p,A}$ already captures how far $p$ drops in general, $\gamma$ is
estimated only from phonemes with $u = 0$ in some pairs and $u = 1$ in others,
which is why only those phonemes contribute.

\textit{The edit experiment.} We fit the same regression to the edit of
Section~\ref{sec:mechanism}, which takes monolingual speech in language $A$
and moves the model's language estimate toward another language $B$. Here
``before'' is the unedited output and ``after'' is the output edited toward
$B$, and $u = 1$ if $B$ lacks $p$ by the training-reference definition of
Appendix~\ref{app:subspace}. Because each of the 16 languages is edited toward
the others, many more phonemes are unshared with some targets and shared with
others: 262 to 300  contribute to the comparison.

Table~\ref{tab:samephoneme} reports $\gamma$ and $e^{\gamma}$ for both
settings, with 95\% bootstrap intervals, refitting the regression on every
resample. We resample the units whose outputs are correlated: clips on Spliced
FLEURS; for the edit, speakers
in the phone recognizers, and utterances in the TTS model, which always speaks
in the same reference voice.

        \begin{table}[t]
        \centering
        \small
        \begin{tabular}{llcc}
        \toprule
        Setting & Model & $\gamma$ (log-odds) & Odds ratio $e^{\gamma}$ \\
        \midrule
        Spliced FLEURS & \powsm{}, CTC head & 0.77 [0.61, 0.97] & 2.2 [1.8, 2.6] \\
         & \powsm{}, attn.\ decoder & 0.84 [0.66, 1.04] & 2.3 [1.9, 2.8] \\
         & \pxeus{} & 0.27 [0.10, 0.47] & 1.3 [1.1, 1.6] \\
         & TTS (CosyVoice) & 0.10 [$-$0.02, 0.22] & 1.1 [0.98, 1.3] \\
        \midrule
        Edit toward $B$ & \powsm{}, CTC head & 2.28 [2.07, 2.53] & 9.8 [7.9, 12.6] \\
         & \pxeus{} & 2.44 [2.31, 2.64] & 11.5 [10.1, 14.1] \\
         & TTS (CosyVoice) & 0.19 [0.13, 0.26] & 1.2 [1.1, 1.3] \\
        \bottomrule
        \end{tabular}
        \caption{\textbf{The same phoneme next to languages that lack it and languages that have it.} $\gamma$: additional drop in the log-odds of recalling a phoneme when the other language lacks it, with the phoneme and the language pair held fixed; 0 if a phoneme's difficulty alone decided the drop. 95\% bootstrap intervals. The TTS interval on Spliced FLEURS includes 0.}
        \label{tab:samephoneme}
        \end{table}

\paragraph{How far a replaced phoneme moves.}
\label{app:subdistance}
Section~\ref{sec:codeswitch-results} reports that unshared phonemes are replaced by distant phonemes of the other language. To measure this, we take each unshared reference position, find the phoneme aligned to it in the transcript (using the same recall edit distance alignment), and keep substitutions by a phoneme of the other language. We keep only the substitutions that the other language adds: the other language splice minus the same language splice on Spliced FLEURS, and \textsc{tog} minus \textsc{sep} on SwitchLingua. We measure distance with PanPhon's weighted feature edit distance~\citep{mortensen2016panphon}, where a voicing change costs 0.25 and /\ipa{\*r}/ to /\ipa{K}/ costs 6.9. We compare it with the distance each substitution would have if it had gone to the most similar phoneme of the other language (Table~\ref{tab:subdistance}; 95\% bootstrap intervals over clips on Spliced FLEURS and conversations on SwitchLingua).

The added substitutions are 3.0 to 4.3 times as far as the most similar phoneme, and only 19\% to 42\% of them are that most similar phoneme. For example, English /\ipa{\*r}/ next to French becomes /\ipa{K}/ (6.9) rather than /j/ (1.6), and English /w/ next to German becomes /v/ (6.25) rather than /j/ (1.5). If the substitutions only reflected small articulatory conversions, where two languages write a similar sound with different symbols, they would go to the most similar phoneme.

There are two caveats to this experiment. First, deletions have no distance and are excluded; they are 8\% to 35\% of the added misses. PanPhon has no feature separating the tap from the trill (distance 0). Secondly, substitutions between them are 4\% to 6\% of the phone recognizers' added substitutions on Spliced FLEURS, none of the TTS model's, and impossible on SwitchLingua, where English has neither.

        \begin{table}[t]
        \centering
        \small
        \begin{tabular}{llccc}
        \toprule
        Dataset & Model & Added substitutions & Most similar phoneme & Share \\
        \midrule
        Spliced FLEURS & \powsm{}, CTC head & 2.64 [2.42, 2.87] & 0.68 & 0.29 \\
         & \powsm{}, attn.\ decoder & 2.86 [2.61, 3.10] & 0.68 & 0.25 \\
         & \pxeus{} & 1.78 [1.46, 2.11] & 0.59 & 0.42 \\
         & TTS (CosyVoice) & 2.47 [2.18, 2.75] & 0.79 & 0.35 \\
        \midrule
        SwitchLingua & \powsm{}, attn.\ decoder & 3.36 [3.03, 3.72] & 0.77 & 0.19 \\
         & \pxeus{} & 2.76 [2.45, 3.10] & 0.70 & 0.29 \\
         & TTS (CosyVoice) & 2.76 [2.36, 3.15] & 0.84 & 0.28 \\
        \bottomrule
        \end{tabular}
        \caption{\textbf{Replacements are far from the phonemes they replace.}
    Added substitutions: mean PanPhon feature distance between an unshared phoneme and the other language's phoneme that replaces it, over the replacements the other language adds (95\% bootstrap intervals).
    Most similar phoneme: the same mean if each had been replaced by the closest phoneme of the other language.
    Share: fraction of replacements that are that closest phoneme.}
        \label{tab:subdistance}
        \end{table}
    
    \section{Unseen languages: detailed results}
    \label{app:lowres}
    
    All numbers are unweighted means over the 43 languages, with 
    95\% bootstrap intervals from clips resampled within language.
    
    \paragraph{Recall by confidence in unseen languages.} From the lowest to the 
    highest third of language estimate confidence, 
    unshared phoneme recall falls by $0.053$ on POWSM and $0.060$ on 
    PhoneticXeus (both intervals exclude zero). Shared phoneme recall 
    rises slightly ($+0.015$ and $+0.029$),  which a general drop in audio quality would not produce. The effect holds when restricted to consonants 
    only ($-0.054$ and $-0.067$), addressing the concern that 
    unshared vowels may reflect transcription conventions more than 
    acoustic contrasts. The assigned language is Thai for 28--29 of the 43 languages, which likely reflects the probe's default for unfamiliar speech more than phonological similarity. The probe chooses among only 16 of the recognizers' training languages. A probe over more of them could assign each unseen language more precisely, which we expect would increase the interference we measure.
    
    \paragraph{Replacements.} The share of unshared phoneme positions 
    written as a phoneme from the assigned language's inventory rises 
    from 0.57 (lowest tercile) to 0.63 (highest) on POWSM 
    ($+0.054$) and from 0.58 to 0.61 on PhoneticXeus ($+0.033$).
    
    \paragraph{WLE on unseen languages.} WLE does not reduce interference 
    on DoReCo because the basis contains none of the 
    languages in DoReCo, and Appendix~\ref{app:langset} shows that a basis 
    missing both languages of a pair repairs much less than the full basis. 
    Additionally, on monolingual input the windowed and full clip 
    language estimates largely agree, so the edit has little to 
    correct. The sliding window also fails here and additionally 
    degrades shared phoneme recall by 0.02--0.03.

    \section{Language probe by layer}
    \label{app:probe}
    
\begin{figure}[t]
        \centering
        \includegraphics[width=\linewidth]{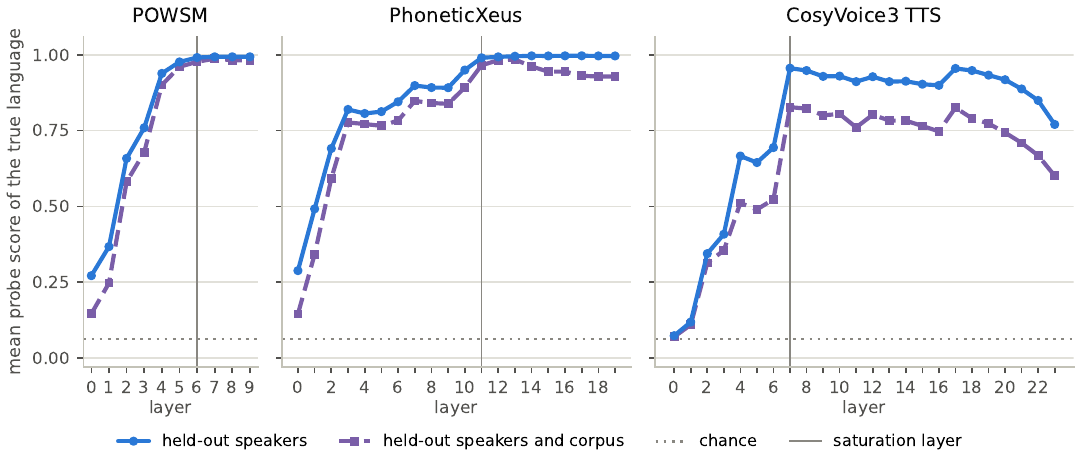}
        \caption{\textbf{Language probe accuracy by layer}, on held-out 
        monolingual speech. \textbf{Held-out speakers}: 
        5,800 utterances from 1,112 unseen speakers. \textbf{Held-out 
        speakers and corpus}: each point is from a probe trained 
        without that language's corpus, so neither the speakers nor 
        the recording pipeline have been seen. The vertical line 
        marks the first layer within 0.01 of the best. For the TTS, 
        the probe reads the language model at each speech token 
        position, with the utterance's own speech tokens as input. 
        We read language estimates at layer 8 (\powsm{}), layer 17 
        (\pxeus{}) and block 7 of the TTS language model.}
        \label{fig:probe}
    \end{figure}

Figure~\ref{fig:probe} shows the language probe's accuracy at each layer,
which we use to choose the layer for all language estimates in the paper.
We evaluate on 5,800 held-out IPAPack++ utterances from unseen speakers
(Appendix~\ref{app:exp_details}).\footnote{FLEURS has no speaker labels,
so we use its speaker-disjoint test split.} Predictions are averaged over
frames, and accuracy is averaged over the 16 languages, so chance is $1/16$.
In both recognizers, accuracy plateaus by mid-network and barely changes
within one or two layers of our choice. It stays high even when the probe
has never seen the target language's corpus, so it is not picking up
recording artifacts. The TTS behaves differently. Accuracy peaks sharply
at block 7 (0.956, up from 0.694 at block 6) and is more sensitive to an
unseen corpus, dropping by 0.129 compared with 0.011 for \powsm{} and 0.066
for \pxeus{}. We therefore take language estimates from layer 8 of
\powsm{}, layer 17 of \pxeus{}, and block 7 of the TTS language model.

        \section{Language subspace additional details}
        \label{app:subspace}
    
        \paragraph{Construction.} At each encoder block, we compute the mean hidden representation of each of the $K=16$ languages over the probe's training frames. These 16 centroids, after centering, span at most $K - 1 = 15$ directions. We whiten by the average within-language covariance $\Sigma_w$ (with a ridge $\lambda = 10^{-3}\,\mathrm{tr}(\Sigma_w)/D$, where $D$ is the hidden dimension) before taking the top-$k$ singular directions, which gives the projection $\mathrm{proj}_{\mathrm{lang}}$ used in Equation~\ref{eq:edit}. The whitening ensures the edit moves along directions that separate languages after accounting for within-language variance, rather than just the directions of largest raw spread. We also tested an unwhitened variant (orthogonal projection onto the centered means directly); it removes slightly more unshared phonemes but degrades shared phonemes more, so we use the whitened version.
    
        \paragraph{Evaluation data.} We evaluate on held-out monolingual speech from IPAPack++ (FLEURS, Common Voice, and audiobook corpora) across all 16 languages, with no speaker overlap with the probe's training set. Each utterance in language $A$ is paired with a target language $B$ and a third language $C$ (238 of the 240 ordered pairs occur).
        
        \paragraph{Scoring.} A phoneme of $A$ is unshared with $B$ if $A$'s training references contain it at least three times and $B$'s never. This differs from the eSpeak-based inventories used in Section~\ref{sec:bleed-results}. Comparison uses exact match after removing length, nasalization and non-syllabic marks. $\Delta$ is the per-utterance drop in recall from the unedited output, averaged within each language and then across the 16 languages. Intervals are 95\% bootstrap over speaker groups within language. About 21\% of the unshared positions differ from the target language only in notation: affricates written with a tie bar where the target language writes the two halves separately (English references write /\ipa{\t{tS}}/ as two phonemes, /\ipa{t}/ and /\ipa{S}/), or phonemes with a diacritic that the target language writes plainly (/\ipa{k\super{h}}/ against /\ipa{k}/). We therefore rescore every edit after reducing each phoneme in the references and the transcripts to its base letters, splitting tie bar affricates and removing diacritics, and recompute the unshared sets from the reduced symbols. This changes $\Delta$ by at most $0.02$ for every edit we report, on both recognizers and the TTS model, and by at most $0.04$ for any edit on either half of the speakers. The reduced reading also removes contrasts that diacritics carry, such as aspiration in Hindi, Korean and Thai, so we keep the unreduced reading as the main one.

        \paragraph{Dev/test split.} Held-out speakers are split in two.  The start block and the whitening were chosen on the development half (2,906 utterances), and the rank as described under hyperparameter selection below; all reported numbers are from the test half (2,890 utterances). Across the 22 edits shared by both halves, every $\Delta$ agrees within 0.022.

        \paragraph{Hyperparameter selection.} We select the start block where we apply the steering to balance effect size against output quality. The start block is the earliest block whose cumulative edit (through to the final block) reaches 90\% of the maximum effect across all start blocks. With blocks numbered from 0, this rule selects block 6 of \powsm{}'s 9 encoder blocks and block 14 of \pxeus{}'s 19, so the edit is applied at blocks 6 to 8 of \powsm{} and blocks 14 to 18 of \pxeus{}. WLE (Section~\ref{sec:repair}) edits the same blocks. Edits with PER above 0.5 relative to the unedited output are excluded as destructive. The rank is $k = 15$, the full span of the 16 centroids.
        
        \begin{figure}[t]
            \centering
            \includegraphics[width=\linewidth]{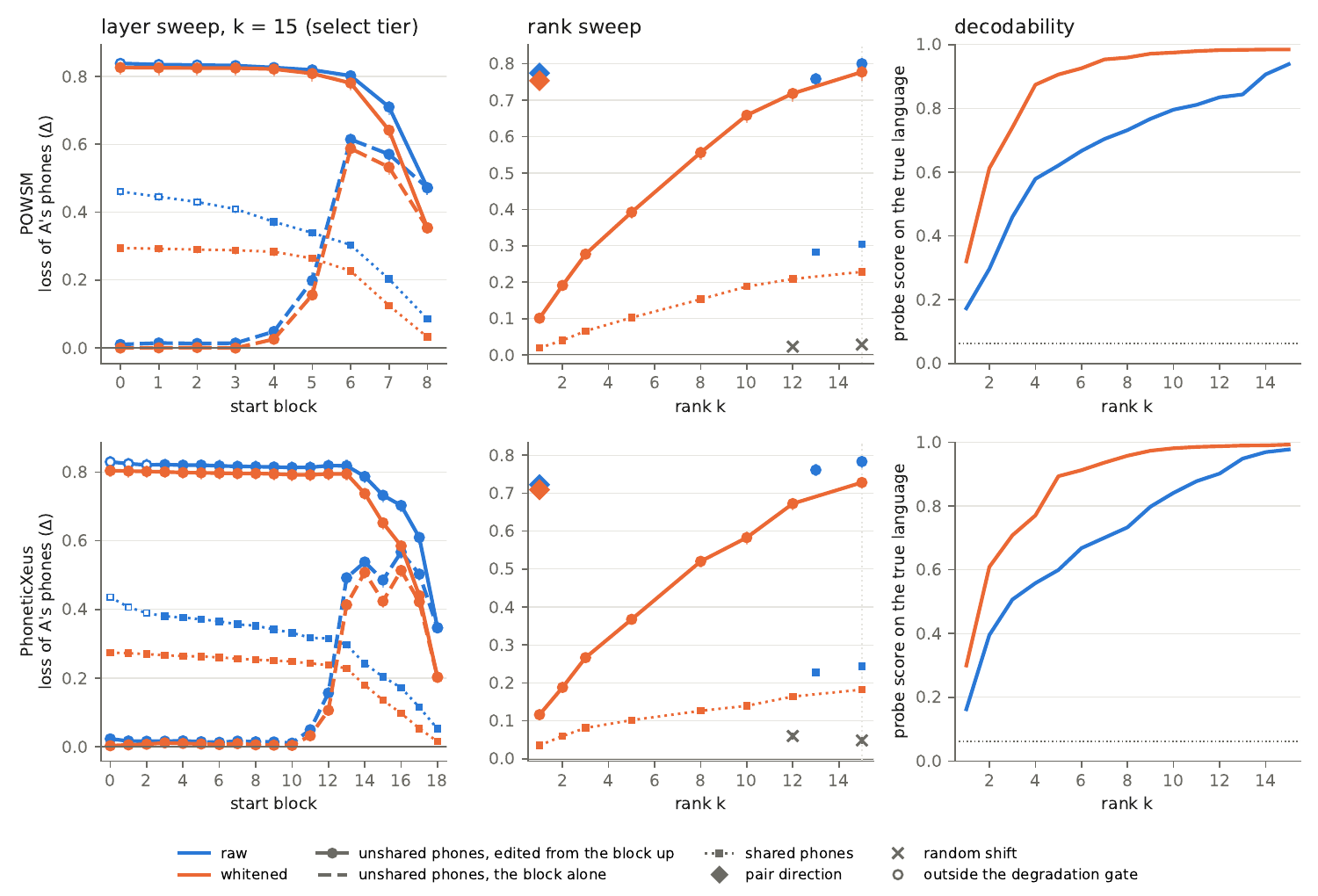}
            \caption{Subspace edit sweeps. \emph{Left:} effect of varying the start block (rank 15, development speakers). \emph{Middle:} effect of varying rank (at the selected start block, test speakers). \emph{Right:} probe accuracy for the true language inside the rank-$k$ subspace (chance dotted). Solid: cumulative edit from start block to last; dashed: single block only. Blue: unwhitened; orange: whitened. See legend for additional 
        markers.}
            \label{fig:sweeps}
        \end{figure}
    
        \paragraph{Rank and depth.} The edit's effect grows with rank and depth (Figure~\ref{fig:sweeps}). Rank 8 reaches 72\% of the rank-15 effect. A probe refitted inside the subspace reads the correct language at $k = 7$ (probe score of $\geq 0.94$), meaning fewer directions suffice for classification than for reproducing interference with steering.
    
        In terms of depth, early blocks contribute little: editing before block 5 (\powsm{}) or block 11 (\pxeus{}) changes recall by at most 0.03. No single block carries more than 75\% of the full stack effect, indicating that the language estimate is recomputed across blocks. Starting the edit at block 0 adds only 0.05--0.07 to the unshared phoneme loss but increases shared phoneme damage by a similar amount.
    
        \paragraph{Controls.} Table~\ref{tab:steering-detail} gives full numbers. Key findings: (1) A random subspace of equal rank shifted by the same amount removes at most 0.06 of unshared phonemes, confirming the effect is specific to the language directions. (2) Erasing language information (setting all frames to the mean of all 16 centroids) removes roughly half the unshared phonemes without imposing any particular language. (3) The phonemes that appear depend on the target language. We count the output phonemes that $B$ uses and neither $A$ nor $C$ does, per 100 output phonemes. The edit toward $B$ raises this count by 7.3 (\powsm{}), 4.6 (\pxeus{}) and 0.9 (TTS), from 0.02 and 0.04 in the recognizers' unedited output. The same edit toward a third language $C$ changes it by at most 0.06, with every interval including zero, while it removes as many of $A$'s phonemes unshared with $C$ as the edit toward $B$ removes of those unshared with $B$ (Table~\ref{tab:steering-detail}). For the TTS model the edit toward $C$ was generated with one seed, on 1,515 utterances. (4) A single direction from $A$'s centroid to $B$'s captures 97\% of the full edit's effect, but requires knowing the source language and so cannot be used as a practical repair. (5) The whitened construction is more precise in the phone recognizers: 
        relative to their unedited recall, the shared-to-unshared drop ratio is 0.29 (\powsm{}) and 0.23 (\pxeus{}{}) with whitening, versus 0.37 and 0.29 without.

        \paragraph{TTS model.} The TTS model receives the same held-out utterances as IPA input. We apply the edit of Equation~\ref{eq:edit} during generation at the positions that produce speech tokens, from block 14 to block 22 of its 24, using the whitened subspace of rank 15 built from the TTS model's own training representations. The start block was selected by the same rule as for the phone recognizers. The last block, 23, is left unedited. Its representations pass through a final normalization before we read them, so an edit there would act on different values than those the subspace was fitted on. Output speech is transcribed by \pxeus{} and compared to the input IPA. Each utterance is generated with four random seeds; edited outputs are compared against unedited outputs at the same seed.
    
        \paragraph{Excluded target languages.} Editing toward four target languages (Korean, Zulu, Kinyarwanda, Vietnamese) caused the TTS output to degenerate: the edited audio was 1.4--1.9 times as long as the original (vs.\ at most 1.25 for other targets), and PER rose to 0.51 (vs.\ 0.34) and PFER to 0.23 (vs.\ 0.12), from 0.20 and 0.07 unedited. We report these four separately. Figure~\ref{fig:steer} and Table~\ref{tab:steering-detail} use the remaining 2,071 utterances. On the excluded four, the edit still removes far more unshared than shared phonemes (84\% vs.\ 32\%), but with more overall degradation. We leave open an investigation into why these four behave differently, which is not clearly explained by training data volume or pretraining language coverage.

    \begin{table}[t]
    \centering
    \small
    
    \textbf{(a) POWSM (CTC head)} --- 2{,}890 utterances, 593 speaker groups.
    Unedited recall: .882 (unshared), .892 (shared).
    
    \vspace{0.3em}
    \begin{tabular}{l cc c}
    \toprule
    & \multicolumn{1}{c}{$\Delta_{\text{unshared}}$} 
    & \multicolumn{1}{c}{$\Delta_{\text{shared}}$} 
    & $\Delta_{\text{sh}} / \Delta_{\text{unsh}}$ \\
    \midrule
    Language subspace $\to B$   
      & .777 {\scriptsize[.757,\,.791]} & .229 {\scriptsize[.220,\,.236]} & .29 \\
    Language subspace $\to C$ (against $C$)
      & .777 {\scriptsize[.756,\,.792]} & --- & --- \\
    One direction ($A\negthinspace\to\negthinspace B$) 
      & .753 {\scriptsize[.730,\,.767]} & .228 {\scriptsize[.220,\,.235]} & .30 \\
    Erased to global mean     
      & .417 {\scriptsize[.399,\,.429]} & .109 {\scriptsize[.103,\,.114]} & .26 \\
    \addlinespace
    Random shift (seed 1)      
      & .029 {\scriptsize[.025,\,.032]} & .006 {\scriptsize[.004,\,.009]} & --- \\
    Random shift (seed 2)      
      & .030 {\scriptsize[.025,\,.035]} & .006 {\scriptsize[.005,\,.008]} & --- \\
    \bottomrule
    \end{tabular}
    
    \vspace{1.2em}
    
    \textbf{(b) PhoneticXeus} --- 2{,}890 utterances, 593 speaker groups.
    Unedited recall: .851 (unshared), .907 (shared).
    
    \vspace{0.3em}
    \begin{tabular}{l cc c}
    \toprule
    & \multicolumn{1}{c}{$\Delta_{\text{unshared}}$} 
    & \multicolumn{1}{c}{$\Delta_{\text{shared}}$} 
    & $\Delta_{\text{sh}} / \Delta_{\text{unsh}}$ \\
    \midrule
    Language subspace $\to B$   
      & .728 {\scriptsize[.714,\,.740]} & .182 {\scriptsize[.177,\,.187]} & .25 \\
    Language subspace $\to C$ (against $C$)
      & .725 {\scriptsize[.713,\,.737]} & --- & --- \\
    One direction ($A\negthinspace\to\negthinspace B$) 
      & .709 {\scriptsize[.695,\,.721]} & .179 {\scriptsize[.174,\,.185]} & .25 \\
    Erased to global mean     
      & .397 {\scriptsize[.381,\,.408]} & .089 {\scriptsize[.085,\,.092]} & .22 \\
    \addlinespace
    Random shift (seed 1)      
      & .048 {\scriptsize[.042,\,.054]} & .015 {\scriptsize[.013,\,.017]} & --- \\
    Random shift (seed 2)      
      & .062 {\scriptsize[.054,\,.068]} & .017 {\scriptsize[.014,\,.019]} & --- \\
    \bottomrule
    \end{tabular}
    
    \vspace{1.2em}
    
    \textbf{(c) CosyVoice3 TTS} --- 2{,}071 utterances, 4 decodes each.
    Unedited recall: .671 (unshared), .822 (shared).
    
    \vspace{0.3em}
    \begin{tabular}{l cc c}
    \toprule
    & \multicolumn{1}{c}{$\Delta_{\text{unshared}}$} 
    & \multicolumn{1}{c}{$\Delta_{\text{shared}}$} 
    & $\Delta_{\text{sh}} / \Delta_{\text{unsh}}$ \\
    \midrule
    Language subspace $\to B$   
      & .346 {\scriptsize[.336,\,.357]} & .107 {\scriptsize[.101,\,.112]} & .31 \\
    Language subspace $\to C$ (against $C$)
      & .347 {\scriptsize[.329,\,.365]} & --- & --- \\
    One direction ($A\negthinspace\to\negthinspace B$) 
      & .337 {\scriptsize[.326,\,.348]} & .105 {\scriptsize[.100,\,.110]} & .31 \\
    Erased to global mean     
      & .227 {\scriptsize[.218,\,.237]} & .050 {\scriptsize[.046,\,.053]} & .22 \\
    \addlinespace
    Random shift (seed 1)      
      & .034 {\scriptsize[.028,\,.041]} & .010 {\scriptsize[.007,\,.013]} & --- \\
    Random shift (seed 2)      
      & .025 {\scriptsize[.012,\,.037]} & .006 {\scriptsize[.001,\,.011]} & --- \\
    \bottomrule
    \end{tabular}
    
    \caption{\textbf{Manipulating the language subspace reproduces interference} (full numbers behind 
    Figure~\ref{fig:steer}). $\Delta$: drop in recall from unedited to edited output, paired per utterance.
    $\Delta_{\text{sh}} / \Delta_{\text{unsh}}$: selectivity ratio; lower means the edit
    targets unshared phonemes more precisely.
    CIs are 95\% bootstrap over speaker groups within language.
    The $\to C$ row gives the drop in recall of $A$'s phonemes unshared with $C$; the other two columns are defined against $B$ and are left empty for it, and for the TTS model it uses 1,515 utterances with one decode each.
    Edited transcripts retain 96--98\% of unedited length in all conditions.}
    \label{tab:steering-detail}
    \end{table}

    \section{Language confusion in Whisper}
    \label{app:whisper}
    
     In this section we show that the same language subspace also explains a failure at the level of words: language confusion in ASR, extending beyond phoneme-level models. Whisper large-v3~\citep{radford2022robustspeechrecognitionlargescale} is a widely used ASR model that outputs words. For every 30\,s of audio, its decoder first outputs one language token and then writes the transcript. We call the language of this token the \emph{picked language}. We use the public checkpoint unchanged and let the model pick the language itself (temperature 0, beam size 5).
    
    Whisper outputs words, so it cannot show interference as defined in Section~\ref{sec:bleed-results}. On code-switched speech it shows language confusion~\citep{marchisio2024understanding}: part of the transcript is in the wrong language or missing. We show here that this failure has the same cause as interference, in three steps. First, on code-switched speech Whisper loses the words of the span that is not in the language it picks. Second, its encoder carries a language estimate that shifts with context, as in the phone recognizers. Third, editing the language subspace of the encoder changes the language Whisper picks, and the picked language decides the language of the transcript. Whisper is a useful test for two reasons. It is widely used, so the failure affects ordinary speech recognition. It also states its language choice itself, so the test does not depend on our phoneme scoring.
    
    \paragraph{Word error rate.} We transcribe SwitchLingua and Spliced FLEURS under the \textsc{sep} and \textsc{tog} conditions of Section~\ref{sec:evaluation}. We measure word error rate (WER, lower is better) against the orthographic reference, after removing case and punctuation. In \textsc{tog}, we align the whole transcript to the whole reference and count the errors of each span separately. WER rises when the other language is present (Table~\ref{tab:whisper-wer}): from $0.16$ to $0.26$ on SwitchLingua and from $0.15$ to $0.54$ on Spliced FLEURS. On Spliced FLEURS the rise appears in both positions (first half $+0.42$, second half $+0.36$), in the 200 clips without English ($+0.40$) and in all 15 language pairs.
    
    \paragraph{Which span loses words?} In each \textsc{tog} recording, one span is in the picked language and the other span is not. We leave out the recordings where Whisper picks a third language (14 in each dataset, mostly Urdu for Hindi). The span in the picked language is transcribed almost as well as it is alone. Almost all of the rise in WER comes from the other span (Table~\ref{tab:whisper-wer}, bottom rows). On SwitchLingua, Whisper usually translates the other span into the picked language. On Spliced FLEURS it usually drops it: in 218 of 286 clips it writes nothing for that span.
    
    \begin{table}[t]
    \centering
    \small
    \begin{tabular}{lcc}
    \toprule
     & SwitchLingua & Spliced FLEURS \\
    \midrule
    WER, \textsc{sep} & .160 & .152 \\
    WER, \textsc{tog} & .259 & .540 \\
    $\Delta$ WER & $+.099$ {\scriptsize[.071,\,.126]} & $+.388$ {\scriptsize[.356,\,.423]} \\
    \midrule
    $\Delta$ WER, span in the picked language & $-.010$ {\scriptsize[$-$.027,\,.006]} & $+.045$ {\scriptsize[.007,\,.097]} \\
    $\Delta$ WER, the other span & $+.223$ {\scriptsize[.168,\,.282]} & $+.741$ {\scriptsize[.695,\,.784]} \\
    \bottomrule
    \end{tabular}
    \caption{\textbf{Whisper loses the span that is not in the language it picks.} WER rises when the other language is present. The span in the picked language is transcribed about as well as it is alone, and almost all of the rise comes from the other span, which Whisper translates or drops. WER is the number of word errors divided by the number of reference words, over both spans. $\Delta$ is \textsc{tog} minus \textsc{sep} on the same recordings, with 95\% bootstrap intervals over recordings. As in Section~\ref{sec:evaluation}, \textsc{sep} is the span alone on SwitchLingua, and the span next to another speaker of the same language on Spliced FLEURS. The bottom rows use the 274 recordings and 286 clips where Whisper picks one of the two languages.}
    \label{tab:whisper-wer}
    \end{table}
    
    \paragraph{Language estimate.} We train the probe of Section~\ref{sec:estimate} on Whisper's encoder, with the same utterances, frames and held-out speakers as for \powsm{}. On held-out speakers, the probe score of the true language is $0.99$ at layer 22. We select this layer as in Appendix~\ref{app:probe}.
    
    On SwitchLingua, the probe score of a span's own language is $0.89$ for the span alone and $0.61$ inside the whole recording. This drop of $0.28$ {\scriptsize[.26,\,.30]} is close to the drop in \powsm{} ($0.31$) and \pxeus{} ($0.29$). On Spliced FLEURS the drop is $0.11$ {\scriptsize[.06,\,.21]}, smaller than in \powsm{} ($0.49$) and \pxeus{} ($0.30$). 
    
    \paragraph{Editing the language subspace.} We build the language subspace on Whisper's encoder as in Appendix~\ref{app:subspace}. We apply the edit of Equation~\ref{eq:edit} to the frames that contain audio, from block 21 to the last block (31). The start block is selected on the development speakers as in Appendix~\ref{app:subspace}. Whisper has no language token for Kinyarwanda or Zulu, which leaves 2,345 test utterances in 14 languages. We measure two things. The first is the probability that Whisper gives to the language token of $B$. The second is whether the transcript is written in $B$, according to a text language identifier ~\citep{stahl2026lingua}. Table~\ref{tab:whisper-steer} shows how both change from the unedited model.
    
    Editing the subspace changes the language Whisper picks, and the transcript follows the pick. The edit toward $B$ lowers the probability of $A$ by $0.45$ and raises the probability of $B$ by $0.20$. The identifier's confidence that the transcript is written in $B$ rises by a similar amount ($0.22$). After the edit, Whisper picks $B$ in 28\% of the utterances, and in those utterances it writes the transcript in $B$. Some of these transcripts are translations. Others spell the sounds of $A$ in the script of $B$.

The edit steers Whisper to the target language. The edit toward a third language $C$ lowers the probability of $A$ just as much, but it raises $C$ ($+0.19$) and leaves $B$ unchanged. When we move the frames to the mean of all 16 language centroids, which points at no language, the probability of $A$ falls by less ($0.28$) and $B$ does not gain. A random subspace of the same rank, shifted by the same amount, has almost no effect.

Finally, we give the decoder the language token of $A$ after the edit toward $B$. The transcript then stays in $A$, and its character error rate rises by only $0.02$. The edit therefore changes the language Whisper writes in for the same audio. Together these results show that the language subspace of the encoder determines the language token, and that the language token determines the language of the transcript.
    
    \begin{table}[t]
    \centering
    \small
    \begin{tabular}{lccc}
    \toprule
    Edit & $\Delta P(B)$ & $\Delta$ transcript in $B$ & $\Delta P(A)$ \\
    \midrule
    Toward $B$ & $+.201$ {\scriptsize[.184,\,.212]} & $+.221$ {\scriptsize[.195,\,.238]} & $-.447$ {\scriptsize[$-$.465,\,$-$.427]} \\
    Toward $C$ & $+.007$ {\scriptsize[.005,\,.008]} & $+.002$ {\scriptsize[.000,\,.003]} & $-.449$ {\scriptsize[$-$.468,\,$-$.429]} \\
    Mean of all centroids & $+.008$ {\scriptsize[.006,\,.010]} & $+.003$ {\scriptsize[.001,\,.005]} & $-.276$ {\scriptsize[$-$.298,\,$-$.261]} \\
    Random subspace & $+.001$ {\scriptsize[.000,\,.001]} & $+.000$ {\scriptsize[.000,\,.000]} & $-.013$ {\scriptsize[$-$.019,\,$-$.010]} \\
    \bottomrule
    \end{tabular}
    \caption{\textbf{Whisper transcribes in the language its encoder is steered toward.} Every edit of the language subspace lowers the probability of $A$, and only the edit toward $B$ raises the probability of $B$ and moves the transcript to $B$. $A$ is the language of the utterance, $B$ is the target language and $C$ is a third language. $P(A)$ and $P(B)$ are the probabilities Whisper gives to the language tokens of $A$ and $B$. ``Transcript in $B$'' is the text language identifier's confidence that the transcript is written in $B$. Each entry is the change from the unedited model, where $P(A)$ is $0.963$. Entries are averaged within each source language and then over the 14 languages, with 95\% bootstrap intervals over speaker groups. The edit toward $C$ uses the 2,027 utterances whose $C$ has a language token.}
    \label{tab:whisper-steer}
    \end{table}
    
    \paragraph{Where the confusion arises.} These results highlight the language confusion source. On Spliced FLEURS, the encoder still gives most frames of both halves to their own language when the other language is present (probe score $0.88$). The decoder picks one language for the whole clip, so only one half can be in the picked language. The other half is translated or dropped. When we give the decoder the correct language token, the transcript is nearly correct even after the edit. The encoder's language estimate therefore shifts with context, as in the phone recognizers, and the decoder's single choice turns this shift into the loss of a whole span.
    
        \section{WLE: confidence intervals and random subspace control}
        \label{app:repair}
    
    Any edit of the hidden states at these blocks could reduce interference simply by disrupting the model, so we test whether the repair depends on the language subspace itself. Table~\ref{tab:repair-detail} gives the intervals behind Table~\ref{tab:repair}. The control applies the same edit, at the same blocks and with the same targets, through a random subspace of equal rank.
    
    \begin{table}[t]
    \centering
    \small
    \begin{tabular}{llcc}
    \toprule
    Model & Condition & $r_{\min}^{\textsc{tog}} \uparrow$ & $\Delta$ (interference) $\downarrow$ \\
    \midrule
    \multirow{9}{*}{TTS}
      & No intervention & .410 & .302 {\scriptsize[.265,\,.340]} \\
      & In chunks, no edit & .351 & .361 {\scriptsize[.328,\,.394]} \\
      & \textsc{wle}, fixed window & .513 & .199 {\scriptsize[.175,\,.223]} \\
      & \textsc{wle}, fixed window, random subspace & .332 & .380 {\scriptsize[.350,\,.411]} \\
      & Span by span, no edit & .383 & .329 {\scriptsize[.291,\,.368]} \\
      & \textsc{wle}, span window & .619 & .093 {\scriptsize[.068,\,.120]} \\
      & \textsc{wle}, span window, random subspace & .383 & .330 {\scriptsize[.292,\,.367]} \\
      & Oracle (known span language) & .606 & .106 {\scriptsize[.079,\,.135]} \\
      & Oracle, random subspace & .388 & .324 {\scriptsize[.286,\,.363]} \\
    \midrule
    \multirow{4}{*}{\pxeus{}}
      & No intervention & .138 & .507 {\scriptsize[.462,\,.551]} \\
      & Sliding window (2\,s) & .443 & .202 {\scriptsize[.160,\,.243]} \\
      & \textsc{wle} & .430 & .214 {\scriptsize[.171,\,.257]} \\
      & \textsc{wle}, random subspace & .150 & .494 {\scriptsize[.449,\,.539]} \\
    \midrule
    \multirow{4}{*}{\powsm{}}
      & No intervention & .146 & .448 {\scriptsize[.404,\,.490]} \\
      & Sliding window (2\,s) & .441 & .153 {\scriptsize[.120,\,.187]} \\
      & \textsc{wle} & .306 & .288 {\scriptsize[.249,\,.327]} \\
      & \textsc{wle}, random subspace & .143 & .451 {\scriptsize[.407,\,.494]} \\
    \bottomrule
    \end{tabular}
    \caption{Interference on SwitchLingua with 95\% bootstrap intervals over recordings. In all three models the random subspace leaves interference at the level of the same decoding without the edit.}
    \label{tab:repair-detail}
    \end{table}
    
The random subspace does not reduce interference in any model. In the recognizers it changes $\Delta$ by at most 0.02 on SwitchLingua, and on Spliced FLEURS it stays within 0.02 of no repair (Table~\ref{tab:oracle}). Paired per recording, WLE leaves 0.16 to 0.28 less interference than the random subspace in every model and window, and every interval excludes zero. On monolingual speech, the two edits cost the same: the random subspace also changes PER and PFER by at most 0.15\,pp (Table~\ref{tab:cost-detail}). Both edits perturb the model by the same amount, and only steering through the language subspace removes interference.

In the TTS model, generating in 12-phoneme chunks without the edit slightly increases interference, by 0.059 [0.025, 0.093], and the random subspace does not undo this. The edit removes a further 0.162 [0.131, 0.194] beyond the chunked decode.

The span window outperforms the fixed window on both interference and cost. It leaves 0.106 [0.074, 0.137] less interference and barely lowers shared phoneme recall: relative to generating each span alone, shared recall drops by 0.075 without intervention, 0.086 with the span window and 0.122 with the fixed window. The gap comes from how often each window's target reflects the right language. We label each target by its nearest language centroid at the last edited block. The span window's target is the chunk's own language on 88\% of chunks. The fixed window's target is on only 79\% of chunks whose window stays within one language, and on 65\% of chunks whose window crosses the switch and so mixes both languages.

    \paragraph{TTS details.} The edit is applied at blocks 14 to 22 of the language model at rank 15, the blocks and rank of the steering edit (Appendix~\ref{app:subspace}). The sentence is generated in chunks of about 12 phonemes, each closed at the next word boundary and generated in the full context of the phonemes and speech tokens before it; the first chunk is never edited, since nothing precedes it. The fixed window is the chunk plus about six phonemes on either side, cut at word boundaries. It is generated with no history and no edit, and the mean of its hidden states at each edited block is the target for every position of the chunk. In the span window arm the chunks are the two language spans. Each sentence is generated with four sampling seeds, which are averaged before scoring. The basis and the oracle's language centroids are fitted on hidden states from real audio, with the speech tokens teacher-forced, while the windowed target and the edit act on the model's own generated states.
    
    \begin{table}[t]
    \centering
    \small
    \setlength{\tabcolsep}{4pt}
    \begin{tabular}{llccc}
    \toprule
    Model & Benchmark & Sliding window & \textsc{wle} & \textsc{wle}, random subspace \\
    \midrule
    \multicolumn{5}{l}{$\Delta$PER (pp)} \\
    \midrule
    \multirow{4}{*}{\powsm{}}
    & TIMIT & $+0.96$ {\scriptsize[$+0.85$,\,$+1.07$]} & $+0.07$ {\scriptsize[$+0.02$,\,$+0.11$]} & $+0.01$ {\scriptsize[$-0.02$,\,$+0.04$]} \\
    & L2-ARCTIC & $+1.55$ {\scriptsize[$+1.37$,\,$+1.74$]} & $+0.10$ {\scriptsize[$+0.01$,\,$+0.19$]} & $+0.02$ {\scriptsize[$-0.02$,\,$+0.07$]} \\
    & GMU accent & $+2.14$ {\scriptsize[$+2.00$,\,$+2.29$]} & $-0.32$ {\scriptsize[$-0.39$,\,$-0.26$]} & $-0.04$ {\scriptsize[$-0.07$,\,$-0.01$]} \\
    & DoReCo & $+1.97$ {\scriptsize[$+1.77$,\,$+2.19$]} & $-0.09$ {\scriptsize[$-0.19$,\,$+0.02$]} & $-0.13$ {\scriptsize[$-0.18$,\,$-0.09$]} \\
    \midrule
    \multirow{4}{*}{\pxeus{}}
    & TIMIT & $+1.33$ {\scriptsize[$+1.22$,\,$+1.44$]} & $+0.08$ {\scriptsize[$+0.05$,\,$+0.11$]} & $+0.01$ {\scriptsize[$-0.01$,\,$+0.03$]} \\
    & L2-ARCTIC & $+3.22$ {\scriptsize[$+3.01$,\,$+3.43$]} & $+0.14$ {\scriptsize[$+0.06$,\,$+0.23$]} & $+0.05$ {\scriptsize[$-0.00$,\,$+0.10$]} \\
    & GMU accent & $+2.71$ {\scriptsize[$+2.55$,\,$+2.87$]} & $-0.09$ {\scriptsize[$-0.17$,\,$-0.02$]} & $-0.02$ {\scriptsize[$-0.05$,\,$+0.01$]} \\
    & DoReCo & $+2.08$ {\scriptsize[$+1.88$,\,$+2.30$]} & $-0.22$ {\scriptsize[$-0.32$,\,$-0.13$]} & $-0.15$ {\scriptsize[$-0.19$,\,$-0.10$]} \\
    \midrule
    \multicolumn{5}{l}{$\Delta$PFER (pp)} \\
    \midrule
    \multirow{4}{*}{\powsm{}}
    & TIMIT & $+0.47$ {\scriptsize[$+0.40$,\,$+0.53$]} & $+0.01$ {\scriptsize[$-0.02$,\,$+0.04$]} & $+0.01$ {\scriptsize[$-0.01$,\,$+0.03$]} \\
    & L2-ARCTIC & $+0.31$ {\scriptsize[$+0.20$,\,$+0.42$]} & $-0.00$ {\scriptsize[$-0.06$,\,$+0.05$]} & $+0.00$ {\scriptsize[$-0.03$,\,$+0.04$]} \\
    & GMU accent & $+0.73$ {\scriptsize[$+0.66$,\,$+0.81$]} & $-0.07$ {\scriptsize[$-0.11$,\,$-0.03$]} & $-0.04$ {\scriptsize[$-0.05$,\,$-0.02$]} \\
    & DoReCo & $-0.45$ {\scriptsize[$-0.61$,\,$-0.28$]} & $+0.05$ {\scriptsize[$-0.02$,\,$+0.11$]} & $-0.15$ {\scriptsize[$-0.18$,\,$-0.12$]} \\
    \midrule
    \multirow{4}{*}{\pxeus{}}
    & TIMIT & $+0.49$ {\scriptsize[$+0.43$,\,$+0.55$]} & $+0.03$ {\scriptsize[$+0.01$,\,$+0.04$]} & $+0.01$ {\scriptsize[$-0.01$,\,$+0.02$]} \\
    & L2-ARCTIC & $+0.50$ {\scriptsize[$+0.40$,\,$+0.61$]} & $+0.03$ {\scriptsize[$-0.01$,\,$+0.08$]} & $-0.02$ {\scriptsize[$-0.05$,\,$+0.02$]} \\
    & GMU accent & $+1.01$ {\scriptsize[$+0.94$,\,$+1.08$]} & $-0.04$ {\scriptsize[$-0.07$,\,$-0.00$]} & $-0.01$ {\scriptsize[$-0.02$,\,$+0.01$]} \\
    & DoReCo & $+1.06$ {\scriptsize[$+0.94$,\,$+1.20$]} & $+0.15$ {\scriptsize[$+0.11$,\,$+0.20$]} & $+0.05$ {\scriptsize[$+0.03$,\,$+0.08$]} \\
    \bottomrule
    \end{tabular}
    \caption{$\Delta$PER (top) and $\Delta$PFER (bottom), in percentage points, behind Table~\ref{tab:cost}, with 95\% bootstrap intervals over clips.}
    \label{tab:cost-detail}
    \end{table}

    \paragraph{Examples.} Table~\ref{tab:wle-examples} shows two sentences in which all three models replace unshared phonemes with the other language's counterparts when they process the whole utterance, and WLE restores them. In (a), the French context turns English \rhotic{} into \uvular{} and \textipa{[U]} into \textipa{[y]}. In (b), the English context turns Spanish \textipa{[R]} into \rhotic{}, and the TTS model also turns \textipa{[x]} into \textipa{[h]}. In the TTS model, these substitutions appear in three or four of the four sampling seeds without the edit and in none with WLE. These are the clearest cases on SwitchLingua: in each model, $r_{\min}$ falls by at least 0.5 from \textsc{sep} to \textsc{tog}, and WLE brings it back to within 0.1 of \textsc{sep}. Seven recordings meet both conditions in all three models, and we chose these two from them to cover both directions. Among all recordings with a lower $r_{\min}$ under \textsc{tog} than under \textsc{sep}, WLE recovers at least half of the drop in 79\% for the TTS model, 57\% for \pxeus{} and 43\% for \powsm{}.

    \begin{table}[t]
    \centering
    \small
    \setlength{\tabcolsep}{5pt}
    \begin{tabular}{@{}llllll@{}}
    \toprule
    Model & Word & Reference & \textsc{sep} & \textsc{tog} & WLE \\
    \midrule
    \multicolumn{6}{@{}p{\linewidth}@{}}{(a)~English words in a French--English sentence: \textit{Le Président américain a annoncé une décision importante aujourd'hui. Can you believe he \textbf{revoked} Biden's \textbf{security} clearance?}} \\
    \addlinespace[2pt]
    TTS & \textit{revoked} & \textipa{\textbf{\textturnr{}}IvoUkt} & \textipa{\textturnr{}ivoUkt} & \textipa{\textcolor{red!75!black}{K}ivOk} & \textipa{\textturnr{}voUkt} \\
     & \textit{security} & \textipa{sIkjU\textbf{\textturnr{}}\textbf{\textturnr{}}Iti} & \textipa{sIkjU\textturnr{}\textturnr{}Idi} & \textipa{sikju\textcolor{red!75!black}{K}iti} & \textipa{sIkjU\textturnr{}\textturnr{}Idi} \\
    \addlinespace[3pt]
    \pxeus{} & \textit{revoked} & \textipa{\textbf{\textturnr{}}IvoUkt} & \textipa{\textturnr{}IvoUkt} & \textipa{\textcolor{red!75!black}{K}ivOkt} & \textipa{\textturnr{}ivoUkt} \\
     & \textit{security} & \textipa{sIkjU\textbf{\textturnr{}}\textbf{\textturnr{}}Iti} & \textipa{sIkjU\textturnr{}@ti} & \textipa{sk\textcolor{red!75!black}{y}\textcolor{red!75!black}{K}iti} & \textipa{skjU\textturnr{}iti} \\
    \addlinespace[3pt]
    \powsm{} & \textit{revoked} & \textipa{\textbf{\textturnr{}}IvoUkt} & \textipa{\textturnr{}IvoUktu} & \textipa{\textcolor{red!75!black}{K}@vOkt} & \textipa{\textturnr{}@vokt} \\
     & \textit{security} & \textipa{sIkjU\textbf{\textturnr{}}\textbf{\textturnr{}}Iti} & \textipa{sIkjU\textturnr{}@ti} & \textipa{sk\textcolor{red!75!black}{y}\textcolor{red!75!black}{K}iti} & \textipa{sIkU\textturnr{}@ti} \\
    \midrule
    \multicolumn{6}{@{}p{\linewidth}@{}}{(b)~Spanish words in a Spanish--English sentence: \textit{Swindon Borough Council está \textbf{trabajando} en este proyecto. It's a big step toward preserving our paleontological heritage.}} \\
    \addlinespace[2pt]
    TTS & \textit{trabajando} & \textipa{t\textbf{R}aba\textbf{x}ando} & \textipa{tRaBaxando} & \textipa{t\textcolor{red!75!black}{\textturnr{}}\textcolor{red!75!black}{A}p@\textcolor{red!75!black}{h}\textcolor{red!75!black}{\~{A}}ndo\textcolor{red!75!black}{U}} & \textipa{tRabaxando} \\
    \addlinespace[3pt]
    \pxeus{} & \textit{trabajando} & \textipa{t\textbf{R}aba\textbf{x}ando} & \textipa{trabaxando} & \textipa{t\textcolor{red!75!black}{\textturnr{}}@b@\textcolor{red!75!black}{\~{\ae{}}}ndo\textcolor{red!75!black}{U}} & \textipa{trabaxando} \\
    \addlinespace[3pt]
    \powsm{} & \textit{trabajando} & \textipa{t\textbf{R}aba\textbf{x}ando} & \textipa{traBaxando} & \textipa{t\textcolor{red!75!black}{\textturnr{}}@b@k\textcolor{red!75!black}{\~{\ae{}}}ndo\textcolor{red!75!black}{U}} & \textipa{tRabaxando} \\
    \bottomrule
    \end{tabular}
    \caption{\textbf{Examples of WLE.} Each model's transcript of the words in bold when it processes each span alone (\textsc{sep}), the whole utterance (\textsc{tog}), and the whole utterance with WLE. For the TTS model, WLE uses the span window, and the transcript is \pxeus{}'s reading of the synthesized span (first sampling seed). In the reference, the unshared phonemes of the word's own language are in bold. In the transcripts, phonemes that only the other language uses are in \textcolor{red!75!black}{red}.}
    \label{tab:wle-examples}
    \end{table}

        \section{Window width}
        \label{app:width}

    Table~\ref{tab:width} repeats the WLE repair with windows of 1, 2, 3 and 4\,s, each with a hop of half its width, on both recognizers and both datasets. WLE reduces interference at every width, and the random subspace leaves it at its unrepaired level at every width ($\Delta$ within 0.02 of no intervention). No single width gives the least interference everywhere. Under WLE, \pxeus{} on SwitchLingua does best at 1\,s and \powsm{} on Spliced FLEURS does best at 3--4\,s, while the other two settings change little between 2 and 3\,s. We report 2\,s throughout.

    The monolingual cost of WLE stays small at the widths we measured. On the benchmarks of Table~\ref{tab:cost-detail}, WLE changes PER by $-0.48$ to $+0.76$\,pp at 1\,s and by $-0.40$ to $+0.18$\,pp at 3\,s (the latter on clips of at least 4.5\,s), and PFER by $-0.08$ to $+0.40$\,pp at 1\,s and by $-0.08$ to $+0.13$\,pp at 3\,s. The sliding window degrades PER by 4.9 to 13.5\,pp at 1\,s and by 0.4 to 2.7\,pp at 3\,s. It degrades PFER by 1.9 to 7.0\,pp at 1\,s, and at 3\,s it changes PFER by $-0.9$ to $+0.6$\,pp. A narrow window is therefore only usable when it is restricted to the language subspace.

    \begin{table}[t]
    \centering
    \small
    \setlength{\tabcolsep}{4pt}
    \begin{tabular}{lcccc}
    \toprule
    Condition & 1\,s & 2\,s & 3\,s & 4\,s \\
    \midrule
    \multicolumn{5}{@{}l}{\textit{\pxeus{}, SwitchLingua (no intervention: .507)}} \\
      Sliding window & .223 {\scriptsize[.188,\,.260]} & \textbf{.202} {\scriptsize[.160,\,.243]} & .207 {\scriptsize[.163,\,.250]} & .246 {\scriptsize[.203,\,.290]} \\
      \textsc{wle} & \textbf{.185} {\scriptsize[.146,\,.222]} & .214 {\scriptsize[.171,\,.257]} & .259 {\scriptsize[.216,\,.304]} & .306 {\scriptsize[.263,\,.349]} \\
    \midrule
    \multicolumn{5}{@{}l}{\textit{\powsm{}, SwitchLingua (no intervention: .448)}} \\
      Sliding window & .245 {\scriptsize[.203,\,.287]} & \textbf{.153} {\scriptsize[.120,\,.187]} & .184 {\scriptsize[.151,\,.218]} & .211 {\scriptsize[.171,\,.251]} \\
      \textsc{wle} & .321 {\scriptsize[.282,\,.359]} & \textbf{.288} {\scriptsize[.249,\,.327]} & .294 {\scriptsize[.255,\,.333]} & .330 {\scriptsize[.289,\,.371]} \\
    \midrule
    \multicolumn{5}{@{}l}{\textit{\pxeus{}, Spliced FLEURS (no intervention: .186)}} \\
      Sliding window & .108 {\scriptsize[.082,\,.135]} & .037 {\scriptsize[.016,\,.059]} & .039 {\scriptsize[.017,\,.062]} & \textbf{.022} {\scriptsize[.001,\,.042]} \\
      \textsc{wle} & .075 {\scriptsize[.050,\,.101]} & \textbf{.047} {\scriptsize[.023,\,.073]} & .052 {\scriptsize[.027,\,.077]} & .054 {\scriptsize[.030,\,.079]} \\
    \midrule
    \multicolumn{5}{@{}l}{\textit{\powsm{}, Spliced FLEURS (no intervention: .459)}} \\
      Sliding window & .214 {\scriptsize[.184,\,.243]} & .081 {\scriptsize[.060,\,.102]} & .048 {\scriptsize[.029,\,.068]} & \textbf{.036} {\scriptsize[.017,\,.055]} \\
      \textsc{wle} & .322 {\scriptsize[.293,\,.351]} & .234 {\scriptsize[.205,\,.264]} & .198 {\scriptsize[.170,\,.227]} & \textbf{.185} {\scriptsize[.158,\,.215]} \\
    \bottomrule
    \end{tabular}
    \caption{Interference $\Delta$ ($\downarrow$) by window width, with 95\% bootstrap intervals over recordings or clips. The unrepaired $\Delta$ is in parentheses, and the lowest value in each row is shown in \textbf{bold}.}
    \label{tab:width}
    \end{table}

        \section{WLE with different language subsets}
        \label{app:langset}
        
        \paragraph{Setup.} We test whether WLE requires all 16 
        training languages in its basis. We refit the subspace 
        (Appendix~\ref{app:subspace}) on various subsets of the 16 
        languages and measure how much interference each basis removes, 
        expressed as a share of the 2\,s sliding window's reduction 
        (1.0 = matches the window, 0.0 = no effect). 
        Table~\ref{tab:langset} shows results across both datasets 
        and both recognizers.

    \begin{table}[t]
    \centering
    \small
    \setlength{\tabcolsep}{3pt}
    \begin{tabular}{lcccc}
    \toprule
     & \multicolumn{2}{c}{SwitchLingua} & \multicolumn{2}{c}{Spliced FLEURS} \\
    \cmidrule(lr){2-3} \cmidrule(lr){4-5}
    Basis & \pxeus{} & \powsm{} & \pxeus{} & \powsm{} \\
    \midrule
    No repair, $\Delta$ & .507 & .448 & .186 & .459 \\
    Sliding window (2\,s), $\Delta$ & .202 & .153 & .037 & .081 \\
    \midrule
    \multicolumn{5}{l}{\emph{Share of the window's reduction}} \\
    All 16 languages, rank 15 (\textsc{wle}) & .96 & .54 & .93 & .60 \\
    7 code-switched languages, rank 6 & .88 & .52 & .92 & .55 \\
    8 Indo-European languages, rank 7 & .92 & .52 & .92 & .50 \\
    All 16 minus one pair language, rank 14 & .83\,--\,.94 & .43\,--\,.53 & .77\,--\,.92 & .51\,--\,.59 \\
    All 16 minus both pair languages, rank 13 & .24\,--\,.34 & .00\,--\,.05 & .50\,--\,.67 & .29\,--\,.35 \\
    8 non-Indo-European languages, rank 7 & .05 & $-.01$ & $-.01$ & .09 \\
    9 languages outside all evaluated pairs, rank 8 & .03 & $-.04$ & $-.03$ & .03 \\
    Random basis, ranks 6--15 & .01\,--\,.04 & $-.03$\,--\,.00 & $-.01$\,--\,.08 & .01\,--\,.05 \\
    \bottomrule
    \end{tabular}
    \caption{Share of the sliding window's interference reduction recovered by
    \textsc{wle} with each basis. Each value is
    $(\Delta_{\text{no repair}} - \Delta_{\textsc{wle}}) \,/\,
     (\Delta_{\text{no repair}} - \Delta_{\text{window}})$.
    The full 16-language basis recovers most of the window's effect.
    Bases missing one language of the evaluated pair show partial repair;
    on SwitchLingua, whose pairs all contain English, bases missing both
    languages recover a quarter to a third of the window's reduction in
    \pxeus{} and no more than a random basis in \powsm{}.}
    \label{tab:langset}
    \end{table}

        \paragraph{Key findings.} The table shows a clear pattern: 
    WLE works when the basis contains at least one language of the 
    pair being evaluated. Specifically:
    
    (1) The full basis and the 7 code-switched languages perform 
    similarly, showing that languages outside the evaluation pairs 
    contribute little.
    
    (2) Removing one language of the pair degrades performance 
    partially. Removing both degrades it further: on SwitchLingua the 
    basis recovers 0.24--0.34 of the window's reduction in \pxeus{} and 
    0.00--0.05 in \powsm{}, the level of a random basis.
    
    (3) Bases containing none of the evaluation languages (the 9 
    uninvolved languages, the 8 non-Indo-European languages on 
    non-Korean pairs) perform no better than random, despite 
    displacing representations by a comparable amount. The subspace 
    directions must include the relevant languages to be effective.
    
    (4) The 16 language centroids sit in near-general position: 
    removing both languages of a pair leaves only 4--15\% of their 
    pairwise direction in the remaining span, comparable to a random 
    subspace of the same rank (1--3\%).

        \section{Oracle repair (known language)}
    \label{app:oracle}
    
    \paragraph{Setup.} WLE estimates the local language from a 
    windowed encoding. As a reference, we test an oracle that 
    knows the true language boundary and sets each frame's subspace 
    coordinates to the centroid of its actual language:
    \begin{equation}
        \mathbf{h} \leftarrow \mathbf{h} 
        - \mathrm{proj}_{\mathrm{lang}}(\mathbf{h}) 
        + \mathrm{proj}_{\mathrm{lang}}(\mathbf{c}_{\ell(t)}),
        \label{eq:oracle}
    \end{equation}
    where $\ell(t)$ is the language of frame $t$. All other settings 
    (basis, rank, blocks, scoring) match WLE in 
    Table~\ref{tab:repair}. A random subspace of equal rank with the 
    same per-frame targets serves as a control. Both run on 
    SwitchLingua and Spliced FLEURS with both recognizers. In the TTS
    model the oracle sets each span's coordinates to the centroid of
    the span's known language, with the same random subspace control;
    Table~\ref{tab:repair-detail} gives its rows.

    \begin{table}[t]
    \centering
    \small
    \setlength{\tabcolsep}{4pt}
    \begin{tabular}{lcccc}
    \toprule
     & \multicolumn{2}{c}{SwitchLingua} & \multicolumn{2}{c}{Spliced FLEURS} \\
    \cmidrule(lr){2-3} \cmidrule(lr){4-5}
    Condition & \pxeus{} & \powsm{} & \pxeus{} & \powsm{} \\
    \midrule
    No intervention
      & .507 {\scriptsize[.462,\,.551]} & .448 {\scriptsize[.404,\,.490]}
      & .186 {\scriptsize[.156,\,.218]} & .459 {\scriptsize[.425,\,.494]} \\
    Sliding window (2\,s)
      & .202 {\scriptsize[.160,\,.243]} & .153 {\scriptsize[.120,\,.187]}
      & .037 {\scriptsize[.016,\,.059]} & .081 {\scriptsize[.060,\,.102]} \\
    \textsc{wle}
      & .214 {\scriptsize[.171,\,.257]} & .288 {\scriptsize[.249,\,.327]}
      & .047 {\scriptsize[.023,\,.073]} & .234 {\scriptsize[.205,\,.264]} \\
    \textsc{wle}, random subspace
      & .494 {\scriptsize[.449,\,.539]} & .451 {\scriptsize[.407,\,.494]}
      & .178 {\scriptsize[.148,\,.211]} & .440 {\scriptsize[.406,\,.475]} \\
    Oracle
      & .108 {\scriptsize[.073,\,.144]} & .121 {\scriptsize[.080,\,.162]}
      & .028 {\scriptsize[.005,\,.050]} & .143 {\scriptsize[.118,\,.169]} \\
    Oracle, random subspace
      & .505 {\scriptsize[.460,\,.549]} & .452 {\scriptsize[.409,\,.493]}
      & .185 {\scriptsize[.155,\,.216]} & .448 {\scriptsize[.414,\,.483]} \\
    \bottomrule
    \end{tabular}
    \caption{Interference $\Delta$ (lower is better) with 95\% intervals.
    The oracle reduces interference beyond \textsc{wle} on all four cells, and
    beyond the sliding window on three. The same edits through a
    random subspace have no effect, confirming the reduction depends
    on the language subspace.}
    \label{tab:oracle}
    \end{table}

\paragraph{Results.} The oracle removes more interference than WLE in all
four cells, with the largest gap on POWSM (Table~\ref{tab:oracle}). It does
not remove all of it: $\Delta$ stays between 0.108 and 0.143, except for
\pxeus{} on Spliced FLEURS (0.028). To check whether the remaining
interference is phoneme-specific, we subtract the shared-phoneme drop from
the unshared-phoneme drop (exact match) for each recording or clip
(Table~\ref{tab:bleeding-detail}). On SwitchLingua, this gap is 0.178
(\pxeus{}) and 0.230 (\powsm) with no intervention and 0.058 and 0.144 under
WLE. Under the oracle it is close to zero (0.002 and 0.011, both CIs
include zero), so the remaining interference affects shared and unshared
phonemes equally. On Spliced FLEURS, the gap also disappears for \pxeus{}
(0.018 [$-0.005$, 0.039]). \powsm, however, keeps 0.086 of its 0.285, so on
POWSM these 15 dimensions do not explain the whole effect. In the TTS
model, the oracle and WLE with the span window leave similar interference
($\Delta = 0.106$ vs.\ $0.093$; paired difference $-0.013$
[$-0.031$, $0.006$]). The two methods differ only in their target, so the
language estimate the model forms from the span alone works as well as the
true language's centroid.

    \paragraph{Caveats.} The oracle edit is not a fair comparison 
    with WLE, for two reasons. First, it requires knowing the 
    language boundary. Second, the oracle subtracts a training corpus 
    centroid from test audio, so the edit also moves each frame 
    toward the training distribution, not just toward the correct 
    language. WLE avoids this because it subtracts two encodings of 
    the same audio. We therefore treat the oracle as a loose upper 
    bound, not a target for WLE to match. The second reason applies to
    the TTS oracle as well, whose centroids come from real audio while
    the span window's target comes from the model's own generated
    states.

    \paragraph{Language tag as an oracle.} When \powsm{}'s attention
    decoder receives each span's own language tag, it shows much less interference,
    and the remaining interference stays above zero on both sets
    (Table~\ref{tab:powsmtag}). \powsm{} takes one language tag per
    utterance, as the first token of its decoder prompt, so we decode
    each span alone under its own language's tag and the whole recording
    once under each of its two languages' tags, and we read each span
    from the decode under its own tag. Like the oracle edit, this needs
    the language of each span and the switch point. On Spliced FLEURS
    the decoder often ends the transcription right after the tag; in
    that case we take the best non-empty hypothesis of the same beam.
    Paired per recording or clip, the tag lowers $\Delta$ by 0.312
    [0.263, 0.360] on SwitchLingua and by 0.197 [0.160, 0.235] on
    Spliced FLEURS. The CTC head of the same encoder receives no tag and
    shows as much interference as the attention decoder under \texttt{<unk>}
    (Table~\ref{tab:bleeding-detail}).

    \begin{table}[t]
    \centering
    \small
    \setlength{\tabcolsep}{4pt}
    \begin{tabular}{lcc}
    \toprule
    Language tag & SwitchLingua & Spliced FLEURS \\
    \midrule
    \texttt{<unk>}
      & .414 {\scriptsize[.371,\,.456]} & .445 {\scriptsize[.408,\,.482]} \\
    Each span's own language
      & .102 {\scriptsize[.070,\,.134]} & .248 {\scriptsize[.213,\,.284]} \\
    Difference
      & .312 {\scriptsize[.263,\,.360]} & .197 {\scriptsize[.160,\,.235]} \\
    \bottomrule
    \end{tabular}
    \caption{\textbf{\powsm{}'s attention decoder with each span's own
    language tag.} Interference $\Delta$ (lower is better), the drop in
    worst span recall of unshared phonemes from \textsc{sep} to
    \textsc{tog}, with 95\% bootstrap intervals, on the recordings (266)
    and clips (274) that both decodes can score. Difference: $\Delta$
    under \texttt{<unk>} minus $\Delta$ under the span's own tag, paired
    per recording or clip.}
    \label{tab:powsmtag}
    \end{table}

    \section{Interference in bilingual speech}
    \label{app:interference}

    \citet{weinreich1953languages} defines interference as the deviations from the norms of either language that occur in the speech of bilinguals as a result of their familiarity with more than one language, and \citet{grosjean2001bilingual} calls the momentary deviations caused by the language not currently spoken \emph{dynamic interferences}. Studies of code-switched speech measure this influence mostly through voice onset time (VOT). Stops near a switch shift toward the other language, by about 28\,ms in a laboratory switching task~\citep{antoniou2011inter} and by 2 to 6\,ms in spontaneous conversation~\citep{piccinini2015voice, fricke2016phonetic}, and a vowel contrast that only Catalan has becomes slightly more similar to Spanish when Spanish words are mixed in~\citep{simonet2014phonetic}. The shift often affects only one of the two languages~\citep{olson2013bilingual, balukas2015spanish}, and some studies find none~\citep{grosjean1994going, gavino2023consequences}. On the perception side, the language context shifts where bilingual listeners place a category boundary~\citep{elman1977perceptual}. In bilinguals, interference is therefore small and gradient, and the models we study show a much larger form of it, replacing whole phonemes in both recognition and synthesis. In multilingual machine learning, \emph{negative interference} names a different effect: languages that share model capacity during training lower each other's performance~\citep{wang2020negative, conneau2020unsupervised}. The interference we study arises at inference, from the context of a single input.

        \end{document}